%% file: main.tex
\documentclass[10pt,twocolumn,letterpaper]{article}

\usepackage[pagenumbers]{cvpr} 

\input{preamble}

\definecolor{cvprblue}{rgb}{0.21,0.49,0.74}
\usepackage[pagebackref,breaklinks,colorlinks,allcolors=cvprblue]{hyperref}

\def\paperID{*****} 
\def\confName{CVPR}
\def\confYear{2026}

\title{A Comprehensive Evaluation of LLM Reasoning: From Single-Model to Multi-Agent Paradigms}

\author{
Yapeng Li$^{1}$, Jiakuo Yu$^{1}$, Zhixin Liu$^{1}$, Xinnan Liu$^{1}$, Jing Yu$^{1}$, Songze Li$^{1}$, Tonghua Su$^{1}$\thanks{Corresponding author} \\[0.5ex]
$^{1}$ Harbin Institute of Technology \\[0.5ex]
\{liyapeng, yujiakuo, Zhixin\_Liu, liuxinnan, yujing,  lisongze\}@stu.hit.edu.cn, \{thsu\}@hit.edu.cn
}

\begin{document}
\maketitle
\input{sec/0_abstract}

\input{sec/1_intro}
\input{sec/2_relat}

\input{sec/3_preli}
\input{sec/4_mineb}
\input{sec/5_exper}
\input{sec/6_deepe}

\input{sec/7_concl}
{
    \small
    \bibliographystyle{ieeenat_fullname}
    \bibliography{main}
}
\input{sec/8_Appendix}

\end{document}

%% file: preamble.tex
\usepackage{booktabs}
\usepackage{multirow}
\usepackage{xcolor}
\usepackage{graphicx}
\usepackage{url}
\usepackage[ruled,vlined]{algorithm2e}
\usepackage{refcount}

%% file: sec/0_abstract.tex
\begin{abstract}

Large Language Models (LLMs) are increasingly deployed as reasoning systems, where reasoning paradigms---such as Chain-of-Thought (CoT) and multi-agent systems (MAS)---play a critical role, yet their relative effectiveness and cost--accuracy trade-offs remain poorly understood.
In this work, we conduct a comprehensive and unified evaluation of reasoning paradigms, spanning direct single-model generation, CoT-augmented single-model reasoning, and representative MAS workflows, characterizing their reasoning performance across a diverse suite of closed-form benchmarks.
Beyond overall performance, we probe role-specific capability demands in MAS using targeted role isolation analyses, and analyze cost--accuracy trade-offs to identify which MAS workflows offer a favorable balance between cost and accuracy, and which incur prohibitive overhead for marginal gains.
We further introduce \textbf{MIMeBench}, a new open-ended benchmark that targets two foundational yet underexplored semantic capabilities---semantic abstraction and contrastive discrimination---thereby providing an alternative evaluation axis beyond closed-form accuracy and enabling fine-grained assessment of semantic competence that is difficult to capture with existing benchmarks.
Our results show that increased structural complexity does not consistently lead to improved reasoning performance, with its benefits being highly dependent on the properties and suitability of the reasoning paradigm itself.
The codes are released at~\url{https://gitcode.com/HIT1920/OpenLLMBench}.
\end{abstract}

%% file: sec/1_intro.tex
\section{Introduction}
\label{sec:intro}

Large Language Models (LLMs)~\cite{zhao2023survey, minaee2024large, kumar2024large} have become a foundational paradigm for general-purpose intelligence, demonstrating strong capabilities in complex reasoning, code synthesis, and scientific problem solving. 
Increasingly, LLMs are instantiated as \emph{reasoning systems} that employ LLMs as core components for structured inference and decision-making, for which reliable operation under practical constraints such as accuracy, budget, and controllability becomes critical.

Within such \emph{reasoning systems}, overall reasoning performance is no longer determined solely by model scale or training data, but increasingly depends on the \emph{reasoning paradigm employed during inference}. Beyond direct single-pass generation, techniques such as CoT reasoning~\cite{wei2022chain} and structured MAS workflows have been widely adopted as system-level paradigms to enhance reasoning quality. In practice, modern reasoning systems often combine multiple paradigms---for example, enabling CoT to strengthen a single model’s step-by-step reasoning, while further leveraging MAS workflows to mitigate CoT errors through external interaction and mutual critique.

Despite this progress, a key unresolved problem persists: the field still lacks a comprehensive and understanding of the cost--accuracy trade-offs of existing reasoning paradigms, as well as how their effectiveness varies across diverse scenarios when deployed as reasoning systems.
Existing studies~\cite{wei2022chain,madaan2023self,du2023improving,chen2024agentverse} tend to focus on proposing new reasoning paradigms over a limited set of benchmarks, leaving several critical issues insufficiently characterized. 
In particular, it remains unclear under what circumstances CoT yields consistent accuracy gains rather than increased verbosity or output variance, and whether MAS provide benefits beyond a strong CoT-enabled single model or instead introduces additional instability.
Moreover, comparisons among these paradigms under realistic budget constraints---where token consumption and multi-call overhead are important considerations---are still lacking.

\begin{figure*}[t]
    \centering
    \includegraphics[width=\linewidth]{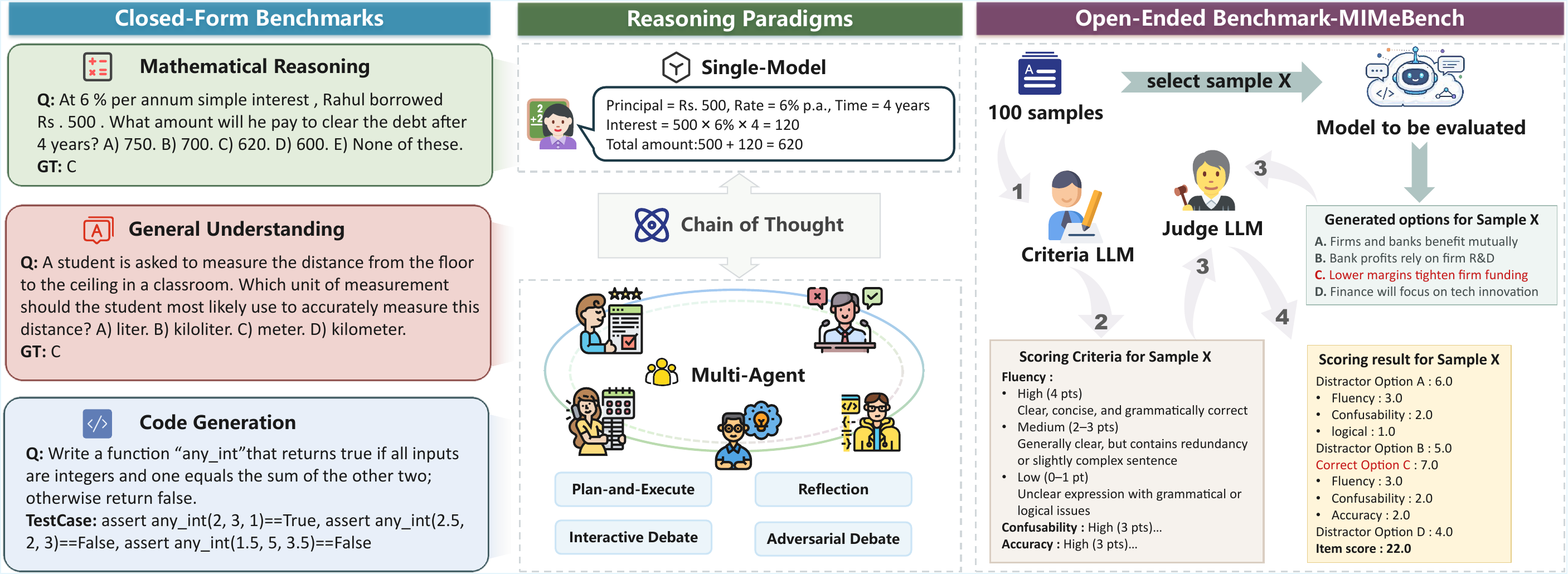}
    \caption{\textbf{Overview of our study.} We evaluate multiple reasoning paradigms under a unified protocol using closed-form benchmarks (\textbf{left}), and complement them with an open-ended benchmark, MIMeBench (\textbf{right}).}
    \label{fig:overview}
\end{figure*}

Motivated by these gaps, we conduct a comprehensive study of reasoning paradigms from \emph{single-model} to \emph{multi-agent}, using an open-weight model \footnote{\label{fn:pangu}\textbf{OpenPangu-Embedded-7B-V1.1.}~\url{https://ai.gitcode.com/ascend-tribe/openpangu-embedded-7b-model}} from the Pangu family~\cite{chen2025pangu} as a representative instance. Concretely, we first establish a rigorous baseline by comparing direct single-model generation against its CoT-enabled counterpart, characterizing CoT's precise impact on correctness and output stability. Building upon this, we systematically evaluate several representative MAS workflows---across a diverse suite of closed-form benchmarks, allowing us to map their effectiveness to specific task domains. We then investigate the interplay between these paradigms by assessing the performance of CoT-augmented MAS, examining whether internal deliberation and external collaboration yield synergistic or diminishing returns. Furthermore, we employ a role isolation protocol to probe the distinct capability demands imposed by different agent roles. Finally, our study concludes with in a fine-grained, cost-aware analysis of evaluated MAS workflows, providing a clear characterization of the accuracy--cost trade-offs to identify which workflows offer a favorable balance of efficiency and reliability, and which incur prohibitive overhead for marginal gains.
However, since our study relies on established closed-form benchmarks, such evaluations are limited to final-answer correctness.
To address this limitation, we introduce \textbf{MIMeBench}, a new open-ended benchmark for main-idea multiple-choice option generation that directly probes two foundational semantic capabilities: \emph{semantic abstraction} and \emph{contrastive discrimination}.
This provides a diagnostic view of the reasoning quality underlying the paradigms we study.
Fig.~\ref{fig:overview} illustrates the overall structure of our study.

Our contributions are summarized as follows:
\begin{itemize}
\item We provide a comprehensive evaluation of reasoning paradigms spanning direct generation, CoT-enabled single-model reasoning, and representative MAS workflows, measuring performance under a unified framework.

\item We introduce MIMeBench, a new open-ended benchmark designed to assess semantic abstraction and contrastive discrimination ability. MIMeBench provides an additional evaluation axis by directly measuring the foundational semantic skills required for robust reasoning.

\item We conduct a detailed analysis of several MAS workflows by examining role-specific capability demands, and by analyzing cost--accuracy trade-offs to determine which workflows offer a favorable accuracy--cost balance and which exhibit diminishing returns.
\end{itemize}

%% file: sec/2_relat.tex
\section{Related Work}
\label{sec:relat}

\subsection{Benchmarks for LLMs}

Benchmarks play a central role in evaluating and comparing large language models, serving as the primary basis for assessing progress across reasoning, knowledge, and code generation.

Existing benchmarks differ substantially in both task formulation and evaluation strategy, and can be broadly grouped into \emph{Closed-Form} Benchmarks, where model outputs are assessed against well-defined ground-truth answers, and \emph{Open-Ended} Benchmarks, where evaluation requires more open-ended judgment.

\paragraph{Closed-Form Benchmarks.}
Closed-form benchmarks span multiple task domains---such as mathematical reasoning, general understanding, and code generation---where they evaluate model outputs using exact answers or deterministic verification procedures.
In mathematical reasoning domain, GSM8K~\cite{cobbe2021gsm8k} serves as a foundational benchmark for grade-school level problems, while AQUA~\cite{ling2017program} targets numerical and algebraic reasoning over text in a multiple-choice setting, and GSM-Hard~\cite{gao2022pal} together with competition-level datasets such as AIME-2024 increase difficulty while preserving answer determinacy.
In general understanding domain, ARC~\cite{allenai:arc} comprises grade-school science questions with Easy and Challenge splits; CommonsenseQA~\cite{talmor-etal-2019-commonsenseqa} targets commonsense knowledge questions; GPQA-Diamond~\cite{rein2024gpqa} is an expert-written 198-question subset spanning biology, chemistry, and physics---all these benchmarks are multiple-choice, with a single correct option as ground truth.
In code generation domain, HumanEval~\cite{chen2021evaluating} adopts a closed-form paradigm by judging functional correctness through unit tests, later strengthened by HumanEval+~\cite{evalplus} with expanded test coverage to improve reliability and reduce false positives.

\paragraph{Open-Ended Benchmarks.}
Open-ended benchmarks target generative tasks where model outputs cannot be evaluated against a single canonical answer, and thus rely more heavily on evaluation procedures.
Traditional automatic metrics such as BLEU~\cite{lifshitz2025multi} and ROUGE~\cite{lin2004rouge} offer scalable scoring but are limited to surface-level overlap and fail to capture semantic correctness or reasoning quality.
To address these limitations, recent benchmarks adopt large language models as judges for open-ended evaluation.
MT-Bench~\cite{zheng2023judging} reports strong agreement between LLM-based judgments and human evaluations.
Building on this paradigm, GPTScore~\cite{fu2024gptscore} and G-Eval~\cite{liu2023g} further formalize evaluation through multi-dimensional criteria and explicit reasoning, improving the reliability and interpretability of open-ended benchmark assessment.


\subsection{LLM-Based Multi-Agent Systems}

Recent advances in LLM reasoning increasingly emphasize structured inference workflows, aiming to improve performance and reliability beyond single-path generation.
Early approaches such as self-ensemble methods~\cite{wangself,yoran2023answering} and iterative self-refinement frameworks~\cite{madaan2023self,shinn2023reflexion,renze2024self} embody this perspective within a single-model setting, by encouraging a single model to generate and aggregate multiple reasoning trajectories or to iteratively revise its outputs through internal feedback.
While effective in improving accuracy, these methods are inherently constrained by single-model introspection and limited exploration, and may degrade when initial reasoning becomes overly confident.

Building upon this workflow-centric perspective, subsequent work generalizes these ideas by externalizing reasoning, critique, and aggregation into explicit interactions among multiple agents. This line of work gives rise to \emph{MAS}, in which distinct agents are explicitly assigned to different roles or stages of the inference workflow, jointly carrying out complex reasoning through coordinated inter-agent interactions~\cite{tran2025multi,li2024survey}.
Representative multi-agent debate frameworks~\cite{du2023improving,liang2024encouraging} show that exchanging conflicting viewpoints can encourage divergent reasoning and improve performance on complex and counter-intuitive tasks. Extensions such as RECONCILE~\cite{chen2024reconcile} and multi-agent verification~\cite{lifshitz2025multi} further highlight the importance of agent diversity, consensus mechanisms, and verification in improving reasoning quality and decision reliability.

In addition, some MAS frameworks move beyond loosely coupled agent interactions and explicitly formalize reasoning workflows as structured, role-based or stage-based decompositions. Systems such as MetaGPT~\cite{hong2023metagpt} and AgentVerse~\cite{chen2024agentverse} decompose complex tasks into coordinated phases, such as planning, execution, and evaluation---enabling fine-grained control and coordination in multi-step problem solving.

%% file: sec/3_preli.tex
\section{Preliminary}
\label{sec:preliminary}

This section introduces the notations and formalizes the difference between \emph{single-model} and \emph{multi-agent reasoning paradigms}. 
We also define the MAS workflows evaluated in this work, which are constructed based on prior work.

\subsection{Notation}
\label{sec:notation}
Let $x$ denote the task input (e.g., a question, a problem statement, or a coding prompt), and $y$ denote the final output (e.g., an answer or a code solution).
We use $\mathcal{M}_\theta$ to denote an LLM with parameters $\theta$.
A decoding procedure induces a conditional distribution:
\begin{equation}
p_\theta(y \mid x) \triangleq \mathcal{M}_\theta(x).
\end{equation}

For any intermediate text artifact (e.g., a plan, critique or an explicit CoT procedure), we denote it by $z$.
A general reasoning process can be viewed as producing a sequence of intermediate artifacts
$\mathbf{z} = (z_1, z_2, \ldots, z_T)$ and then the final output $y$:
\begin{equation}
p_\theta(y, \mathbf{z} \mid x) = \prod_{t=1}^{T} p_\theta(z_t \mid x, z_{<t}) \cdot p_\theta(y \mid x, \mathbf{z}).
\end{equation}

We use $\mathcal{C}(\cdot)$ to denote inference cost (token consumption). For a dialog-style workflow producing messages $\{m_k\}_{k=1}^{K}$, we write
\begin{equation}
\mathcal{C} \triangleq \sum_{k=1}^{K} |m_k|.
\end{equation}
where $|m_k|$ is the number of tokens in message $m_k$.

\subsection{Single-Model Reasoning Paradigm}
\label{sec:single_agent_reason}
We first formalize the \emph{single-model reasoning paradigm}, where a single model instance produces the final answer in one pass:
\begin{equation}
y = f_\theta(x), \quad \text{where } f_\theta(x) \sim p_\theta(y \mid x).
\end{equation}

Optionally, single-model reasoning may generate an explicit CoT procedure $z$:
\begin{equation}
z \sim p_\theta(z \mid x), \quad y \sim p_\theta(y \mid x, z).
\end{equation}
In practice, Pangu-7B supports two inference strategies that can be abstracted as:
(i) \textbf{Direct Response} (\texttt{no\_think}): $y \sim p_\theta(y \mid x)$,
(ii) \textbf{Adaptive Reasoning} (\texttt{auto\_think}): $y \sim p_\theta(y \mid x, z)$ with $z$ generated adaptively.

Accordingly, the inference cost is dominated by a single forward generation, with an optional CoT procedure $z$:
\begin{equation}
y \sim p_\theta(y \mid x, z), \quad \mathcal{C} \approx O(|y| + |z|).
\end{equation}
where $|z|=0$ in the direct-response setting.

\subsection{Multi-Agent Reasoning Paradigm}
\label{sec:multi_agent_reason}
MAS externalize reasoning into explicit interactions among multiple agent instances.
Let there be $N$ agents $\{\mathcal{A}_i\}_{i=1}^{N}$, where each agent $\mathcal{A}_i$ is an instantiation of (possibly the same) base model $\mathcal{M}$ under a role-specific prompt $\pi_i$:
\begin{equation}
\mathcal{A}_i(\cdot) \triangleq \mathcal{M}\big(\pi_i, \cdot \big).
\end{equation}

A general MAS workflow defines (1) a message-passing protocol and (2) a termination rule producing the final output:
\begin{equation}
m_k = g_k\big(x, m_{<k}\big), \qquad y = h\big(x, m_{1:K}\big).
\end{equation}
where $g_k$ specifies which agent speaks at step $k$ and what context it receives, and $h$ aggregates the transcript to form the final prediction.

Compared to single-model, \emph{multi-agent reasoning paradigm} introduces explicit interactive messages, and its inference cost scales with the total length of all generated messages:

\begin{equation}
\mathcal{C} \approx O\Big(\sum_k |m_k| +|y|\Big).
\end{equation}

\subsection{MAS Workflows}
\label{sec:mas_workflows}
\begin{figure}[t]
  \centering
  \includegraphics[width=0.95\linewidth]{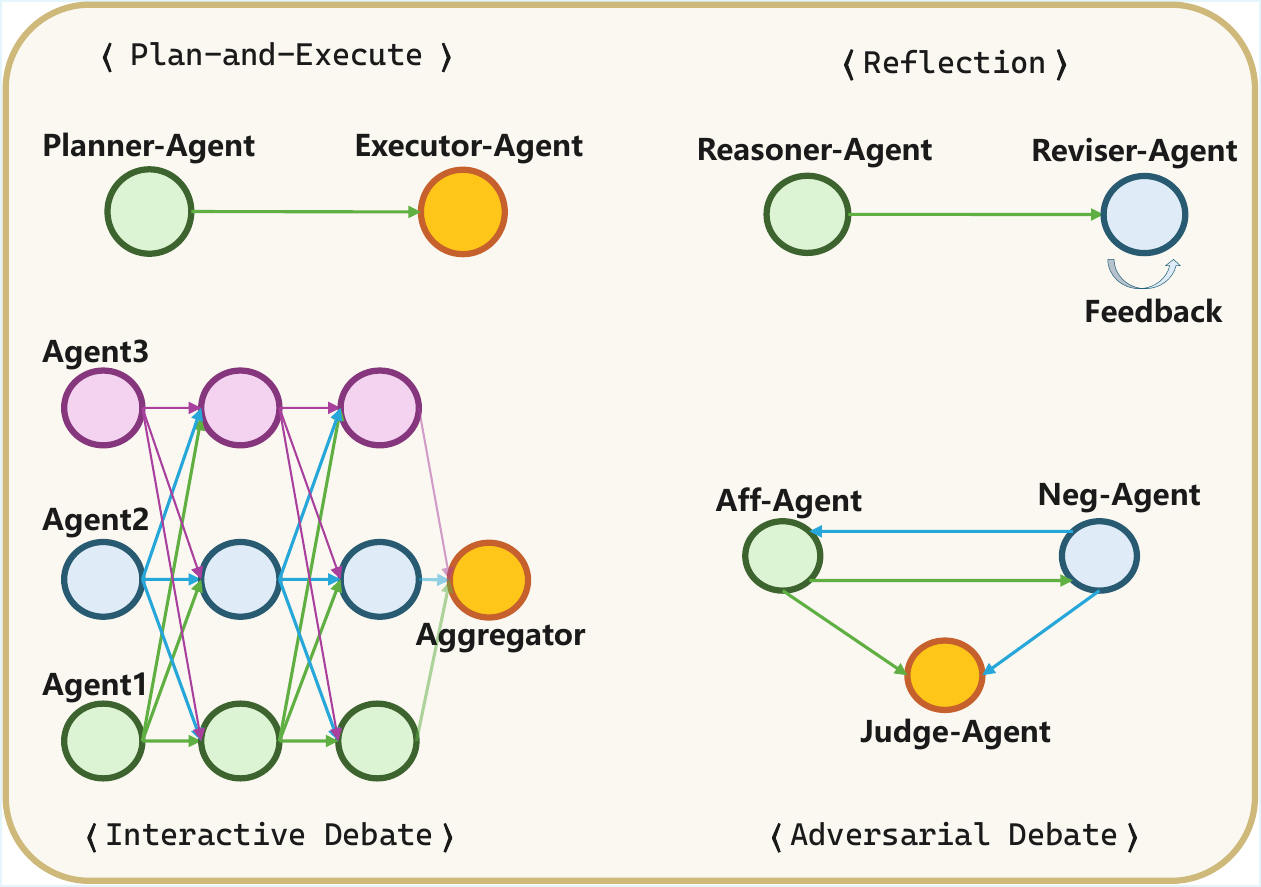}
  \caption{Overview of MAS Workflows.}
  \label{fig:mas-workflow}
\end{figure}

As illustrated in Fig.~\ref{fig:mas-workflow}, we formalize four MAS workflows evaluated in this work---\emph{Plan-and-Execute}, \emph{Reflection}, \emph{Interactive Debate}, and \emph{Adversarial Debate}.

\paragraph{(1) Plan-and-Execute.}
This workflow decomposes problem solving into planning and execution using two agents:
a Planner $\mathcal{A}_{\text{plan}}$ and an Executor $\mathcal{A}_{\text{exec}}$.
First, the Planner generates a plan $z_{\text{plan}}$:
\begin{equation}
z_{\text{plan}} \sim p_{\text{plan}}(z \mid x),
\end{equation}
then the Executor produces the final answer conditioned on the plan:
\begin{equation}
y \sim p_{\text{exec}}(y \mid x, z_{\text{plan}}).
\end{equation}
This design isolates strategic decomposition from instruction-following fidelity.

\paragraph{(2) Reflection.}
This workflow performs iterative correction with two phases.
First, a Reasoner $\mathcal{A}_{\text{rsn}}$ generates an initial solution $y^{(0)}$ (and its rationale).
Then, a Reviser $\mathcal{A}_{\text{rev}}$ first produces an explicit feedback artifact $z_{\text{fee}}$ by critiquing the initial solution, and subsequently generates a revised solution $y^{(1)}$ conditioned on this feedback:
\begin{align}
y^{(0)} &\sim p_{\text{rsn}}(y \mid x), \\
z_{\text{fee}} &\sim p_{\text{rev}}(z \mid x, y^{(0)}), \\
y^{(1)} &\sim p_{\text{rev}}(y \mid x, y^{(0)}, z_{fee}).
\end{align}
We take $y \triangleq y^{(1)}$ as the final output.

\paragraph{(3) Interactive Debate.}
Let there be $N$ peer debaters $\{\mathcal{A}_i\}_{i=1}^{N}$ and an Aggregator $\mathcal{A}_{\text{agg}}$.
Each debater first produces an independent solution:
\begin{equation}
y_i^{(0)} \sim p_i(y \mid x), \quad i=1,\ldots,N.
\end{equation}
For debate rounds $r=1,\ldots,R$, each agent updates its answer conditioned on other agents' synthesized messages $\mathrm{Sync}(\cdot)$:
\begin{equation}
y_i^{(r)} \sim p_i\big(y \mid x, \mathrm{Sync}(y_{-i}^{(r-1)})\big),
\end{equation}
where $y_{-i}^{(r-1)}$ denotes the set of other agents' solutions at round $r-1$.
Finally, $\mathcal{A}_{\text{agg}}$ produces the final output by examining all candidate answers $\{y_i^{(R)}\}_{i=1}^{N}$ and selecting the most frequently occurring one:

\begin{equation}
y = \mathcal{A}_{\text{agg}}\big(\{y_i^{(R)}\}_{i=1}^{N}\big).
\end{equation}

\paragraph{(4) Adversarial Debate.}
This workflow assigns explicit opposing roles: an Affirmative agent $\mathcal{A}_{\text{aff}}$, a Negative agent $\mathcal{A}_{\text{neg}}$, and a Judge $\mathcal{A}_{\text{judge}}$.
The Affirmative proposes an initial solution $y_{\text{aff}}^{(0)}$, and the Negative responds with a counter-solution $y_{\text{neg}}^{(0)}$:
\begin{align}
y_{\text{aff}}^{(0)} &\sim p_{\text{aff}}(y \mid x), \\
y_{\text{neg}}^{(0)} &\sim p_{\text{neg}}(y \mid x, y_{\text{aff}}^{(0)}).
\end{align}
For rebuttal rounds $r=1,\ldots,R$, each side responds to the opponent’s latest message:
\begin{align}
y_{\text{aff}}^{(r)} &\sim p_{\text{aff}}\big(y \mid x, y_{\text{neg}}^{(r-1)}\big),\\
y_{\text{neg}}^{(r)} &\sim p_{\text{neg}}\big(y \mid x, y_{\text{aff}}^{(r)}\big).
\end{align}
The Judge $\mathcal{A}_{\text{judge}}$ then outputs the final decision after reading the complete debate transcript $\mathcal{T}$:

\begin{equation}
\mathcal{T} = \big\{ y_{\text{aff}}^{(r)},\, y_{\text{neg}}^{(r)} \big\}_{r=0}^{R},
\end{equation}

\begin{equation}
y \sim p_{\text{judge}}(y \mid x, \mathcal{T}).
\end{equation}

%% file: sec/4_mineb.tex
\section{MIMeBench}
\label{sec:mimebench}

We introduce \textbf{MIMeBench}, a benchmark for \emph{Main-Idea Multiple-Choice Question (MCQ) Generation}, to evaluate foundational semantic skills that underpin effective reasoning. Unlike closed-form benchmarks, which primarily assess final-answer correctness, MIMeBench directly evaluates a model’s ability to (i) identify the core semantics of a passage and (ii) distinguish between semantically similar yet meaningfully distinct alternatives. These capabilities correspond to \emph{semantic abstraction} and \emph{contrastive discrimination}, respectively.

Rather than assessing only whether a model produces a correct final answer, this open-ended formulation directly measures the quality of two foundational reasoning components---\emph{semantic abstraction} and \emph{contrastive discrimination}---by evaluating how accurately core meaning is extracted and how effectively semantically challenging alternatives are constructed, thereby yielding interpretable signals that help explain and predict performance on complex reasoning tasks.

This section describes the construction of MIMeBench, its dynamic, item-specific evaluation criteria, and the scoring and aggregation protocol used for model assessment.

\subsection{Dataset Construction}
The dataset is compiled from official National Civil Service Examination items and multiple provincial Administrative Aptitude Test (AAT) exams collected over the past five years. We select 100 main-idea summarization samples covering diverse topics and discourse structures. Each item is derived from a real examination question and consists of a passage, a question (typically phrased as \emph{``This passage is intended to illustrate\ldots''}), and four \emph{expert-designed} options as reference, including one correct main-idea option and three distractors. 

Given the passage and prompt, a model is required to generate four new options following the same structure---one correct option and three distractors---mirroring the format and difficulty of authentic examination items. Passage length and difficulty are controlled to reduce bias from extreme cases, while topic diversity is maintained to evaluate contextual generalization. For exam security and compliance reasons, we do not release the original items or full passages.

\subsection{Dynamic Evaluation Criteria}
Unlike closed-form benchmarks with fixed answers or static rubrics, MIMeBench relies on \emph{item-specific evaluation criteria} that capture the semantic structure and distractor logic of each item. This design is motivated by the observation that each item differ substantially in discourse organization, thematic focus, and plausible distractor strategies, making a single global rubric inadequate.

For each benchmark item, a criteria model, denoted as $M_{\text{crit}}$, is prompted to analyze the source passage together with the original reference options from the item. By using these \emph{expert-designed} options, $M_{\text{crit}}$ can derive criteria that align with the experts’ intended interpretation and quality standards for the item. Accordingly, this model is used exclusively to generate item-specific evaluation criteria and is not involved in option generation or scoring.
Based on these information, the model generates two sets of evaluation criteria:
(i) criteria for assessing the correct option, and
(ii) criteria for assessing distractor options.
Within each item, the three distractors are evaluated against the \emph{same} set of distractor criteria to enforce a uniform judging standard, ensuring that the resulting scores are directly comparable across distractors. To reduce stochasticity, three independent sets of criteria are generated for correct options and three for distractors, and scores obtained under these criteria are averaged during aggregation.

\subsection{Scoring Dimensions and Aggregation}
We formalize the scoring process using explicit notation. For a given item, let $o^\star$ denote the correct option generated by the evaluated model, and $\{o_1, o_2, o_3\}$ denote the three generated distractors. Let $\mathcal{C}^\star = \{c^\star_1, c^\star_2, c^\star_3\}$ denote the three independently generated evaluation criteria for the correct option, and $\mathcal{C}^- = \{c^-_1, c^-_2, c^-_3\}$ denote the three evaluation criteria shared by all distractors.

Each correct option is evaluated along three dimensions \emph{fluency}, \emph{confusability}, and \emph{accuracy}---and each distractor along \emph{fluency}, \emph{confusability}, and \emph{logical consistency}, with the scores of the three dimensions summing to a total of 10 points per option. The weighting of these dimensions is fixed across all items.
Here, \emph{fluency} measures grammaticality and readability; \emph{accuracy} measures whether the correct option captures the main idea; for correct options, \emph{confusability} rewards paraphrased expressions that are not trivially anchored by lexical overlap with the source passage(e.g., copying many words), whereas for distractors \emph{confusability} measures how misleading the option is; \emph{logical consistency} checks whether a distractor is internally coherent and not self-contradictory.

For the correct option, the aggregated score is computed as:
\begin{equation}
S^\star = \frac{1}{|\mathcal{C}^\star|} \sum_{k=1}^{|\mathcal{C}^\star|} J(o^\star \mid c^\star_k),
\end{equation}
where $J(\cdot \mid c)$ denotes the judge model scoring an option under criterion $c$.

Similarly, each distractor $o_i$ is scored as:
\begin{equation}
S_i = \frac{1}{|\mathcal{C}^-|} \sum_{k=1}^{|\mathcal{C}^-|} J(o_i \mid c^-_k), \quad i \in \{1,2,3\}.
\end{equation}

The final item-level score is:
\begin{equation}
S_{\text{item}} = S^\star + \sum_{i=1}^{3} S_i.
\end{equation}

Model performance on MIMeBench is reported as the mean item score over the dataset. 
Algorithm~\ref{alg:mimebench_eval} summarizes the full dataset-level evaluation pipeline.

\begin{algorithm}[t]
\caption{MIMeBench evaluation pipeline.}
\label{alg:mimebench_eval}
\small
\KwIn{
Dataset $\mathcal{D}=\{(p^{(n)}, q^{(n)}, \mathcal{R}^{(n)})\}_{n=1}^{N}$ ($N{=}100$), 
evaluated model $M$, 
criteria model $M_{\text{crit}}$, judge model $J$, 
criteria prompts $\pi^\star$ for correct-option criteria and $\pi^-$ for distractor criteria
}
\KwOut{Mean MIMeBench score $\overline{S}$}

\textit{$p$: passage text; $q$: prompt used to elicit $M$ to generate options; $\mathcal{R}$: reference options}\;

$\text{Total} \leftarrow 0$\;

\For{$n \leftarrow 1$ \KwTo $N$}{
    $(p,q,\mathcal{R}) \leftarrow (p^{(n)}, q^{(n)}, \mathcal{R}^{(n)})$\;

    $(o^\star, o_1, o_2, o_3) \leftarrow M(p,q)$\;

    $\mathcal{C}^\star \leftarrow \emptyset$\;
    \For{$k \leftarrow 1$ \KwTo $3$}{
        $c \leftarrow M_{\text{crit}}(p,\mathcal{R}; \pi^\star)$\;
        $\mathcal{C}^\star \leftarrow \mathcal{C}^\star \cup \{c\}$\;
    }

    $\mathcal{C}^- \leftarrow \emptyset$\;
    \For{$k \leftarrow 1$ \KwTo $3$}{
        $c \leftarrow M_{\text{crit}}(p,\mathcal{R}; \pi^-)$\;
        $\mathcal{C}^- \leftarrow \mathcal{C}^- \cup \{c\}$\;
    }

    $S^{(n)} \leftarrow \textsc{ItemScore}(o^\star,o_1,o_2,o_3,\mathcal{C}^\star,\mathcal{C}^-,J)$\;

    $\text{Total} \leftarrow \text{Total} + S^{(n)}$\;
}

$\overline{S} \leftarrow \text{Total} / N$\;
\Return{$\overline{S}$}\;

\vspace{0.5em}
\SetKwProg{Fn}{Function}{:}{end}

\Fn{\textsc{ItemScore}$(o^\star,o_1,o_2,o_3,\mathcal{C}^\star,\mathcal{C}^-,J)$}{
    $S^\star \leftarrow \textsc{AvgCritScore}(o^\star,\mathcal{C}^\star,J)$\;
    $S^- \leftarrow 0$\;
    \For{$i \leftarrow 1$ \KwTo $3$}{
        $S^- \leftarrow S^- + \textsc{AvgCritScore}(o_i,\mathcal{C}^-,J)$\;
    }
    \Return{$S^\star + S^-$}\;
}

\Fn{\textsc{AvgCritScore}$(o,\mathcal{C},J)$}{
    $s \leftarrow 0$\;
    \ForEach{$c \in \mathcal{C}$}{
        $s \leftarrow s + J(o \mid c)$\;
    }
    \Return{$s / |\mathcal{C}|$}\;
}
\end{algorithm}

%% file: sec/5_exper.tex
\section{Experiments}
\label{sec:resul}

This section reports our experimental design and empirical findings. We first describe the evaluation setup, benchmarks, and scoring methodology (Sec.~\ref{sec:ep}). We then present single-model inference results, including cross-model comparisons and the impact of inference strategies (Sec.~\ref{sec:foundational_results}), followed by multi-agent inference results under representative MAS workflows (Sec.~\ref{sec:mas_results}). Finally, we report open-ended evaluation results on MIMeBench.~(Sec.~\ref{sec:mimebench_results}).

\subsection{Experimental Protocol}
\label{sec:ep}
\subsubsection{Setup}
We adopt Pangu-7B \textsuperscript{\hyperref[fn:pangu]{\getrefnumber{fn:pangu}}} model from the Pangu family, which is developed within the Ascend ecosystem. Accordingly, all our evaluation experiments are conducted in an Ascend-based environment, with the models deployed on Ascend 910B NPUs.

To ensure reproducibility and consistency, all evaluated models we used (not only Pangu-7B) are run with the default decoding hyperparameters specified in their open-source configurations (temperature, top\_p, and top\_k). The maximum context length is set to each model's maximum supported embedding length. Except for MIMeBench, all benchmarks are conducted under a unified zero-shot setting without additional prompting or task-specific guidance.

\subsubsection{Benchmarks}
\label{sec:tm}
To comprehensively evaluate the model’s capabilities under diverse reasoning demands, we adopt a suite of closed-form benchmarks, covering both standard evaluations and more rigorous variants. This suite enables a holistic assessment of exact reasoning performance and answer correctness. The specific tasks and their corresponding evaluation metrics are summarized in Table~\ref{tab:foundational_tasks}.

In addition to closed-form benchmarks, we include MIMeBench as an open-ended generation benchmark. The evaluated model is required to generate a set of options for a question, where quality is assessed by semantic adequacy and distractor plausibility rather than exact string matching. Following the protocol in Sec.~\ref{sec:mimebench}, we use a LLM-based judge to score model outputs and report performance using mean scores.

\begin{table}[t]
\centering
\small
\setlength{\tabcolsep}{6pt}
\caption{Selected benchmarks in our work, covering mathematical reasoning, general understanding, and code generation domains (referred to as \textbf{Math}, \textbf{General}, and \textbf{Code} in later analyses), together with an open-ended generation benchmark---MIMeBench.}
\label{tab:foundational_tasks}
\begin{tabular}{llc}
\toprule
\textbf{Domain} & \textbf{Datasets} & \textbf{Metric} \\
\midrule
\multirow{4}{*}{\textbf{Mathematical Reasoning}} 
    & AQUA & \multirow{4}{*}{Accuracy} \\
    & GSM8K & \\
    & GSM-Hard & \\
    & AIME-2024 & \\
\midrule
\multirow{4}{*}{\textbf{General Understanding}}
    & ARC-Easy & \multirow{4}{*}{Accuracy} \\
    & ARC-Challenge & \\
    & CommonsenseQA & \\
    & GPQA-Diamond & \\
\midrule
\multirow{2}{*}{\textbf{Code Generation}}
    & HumanEval & \multirow{2}{*}{Pass@1} \\
    & HumanEval+ & \\
\midrule
\multirow{1}{*}{\textbf{Open-ended Generation}}
    & MIMeBench & Avg. Score \\
\bottomrule
\end{tabular}
\end{table}

\subsubsection{Evaluation Methodology}

To maintain consistency and assessment fidelity for closed-form benchmarks (excluding MIMeBench), we adopt a zero-shot evaluation framework in which Qwen3-32B is used as an automated judge to compare model outputs against ground-truth answers. This framework mitigates parsing errors and standardizes the scoring methodology. We detail the evaluation procedures for different benchmarks below:
\begin{itemize}
    \item \textbf{Math \& General:} For non-coding benchmarks, the model's output and ground truth are fed into Qwen3-32B. The judge performs approximate equivalence checking to ascertain correctness, yielding a binary score of 1 (Correct) or 0 (Incorrect).
    
    \item \textbf{Code:} For coding benchmarks, we primarily rely on a rule-based matching procedure to extract executable code blocks. In cases where the rule-based approach fails to produce a valid extraction, we fall back to using Qwen3-32B as an extractor to isolate the executable code blocks. The extracted blocks are then evaluated against a standard unit test suite: a sample is assigned a score of 1 (Pass) only if it passes all test cases; otherwise, it is assigned 0 (Fail).
\end{itemize}
 For the automated judge, we set the decoding temperature to 0 to reduce stochasticity and promote fair and stable judgments. 


\subsection{Single-Model Inference Results}
\label{sec:foundational_results}

We first establish a model-grounded reference for reasoning performance to situate the subsequent analysis. Adhering to the protocols defined in Sec.~\ref{sec:ep}, we assess Pangu-7B across the selected benchmarks. We benchmark Pangu-7B against contemporary open-weight reasoning models, including the Qwen3 series~\cite{qwen3technicalreport} and the DeepSeek-R1 distilled variants~\cite{deepseekai2025deepseekr1incentivizingreasoningcapability}, and also report its results under both direct-generation and thinking strategies. Together, these results delineate the empirical regime in which our later comparisons are made.

\subsubsection{Comparison with State-of-the-Art Baselines}

We benchmark Pangu-7B (\texttt{auto\_think}) against Qwen3 (8B/14B) and DeepSeek-R1 (Distill-Llama-8B/Qwen3-8B). To ensure a fair comparison, all models are evaluated in their thinking modes.

\begin{table*}[t]
\centering
\normalsize
\setlength{\tabcolsep}{4pt}
\caption{Comparison with state-of-the-art open-weight models.}
\label{tab:cross_model_new}
\begin{tabular}{lll c cc cc}
\toprule
\multirow{2}{*}{\textbf{Domain}} & \multirow{2}{*}{\textbf{Task}} & \multirow{2}{*}{\textbf{Metric}} & \textbf{Pangu} & \multicolumn{2}{c}{\textbf{Qwen3}} & \multicolumn{2}{c}{\textbf{DeepSeek-R1 Distill}} \\
\cmidrule(lr){4-4} \cmidrule(lr){5-6} \cmidrule(lr){7-8}
 & & & \textbf{7B} & \textbf{8B} & \textbf{14B} & \textbf{Llama-8B} & \textbf{Qwen3-8B} \\
\midrule
\multirow{2}{*}{Math} & GSM8K & Acc. & 94.54 & \textbf{97.19} & 97.12 & 82.26 & 95.53 \\
 & AIME-2024 & Acc. & \textbf{86.67} & 80.00 & 80.00 & 53.33 & 80.00 \\
\midrule
\multirow{2}{*}{General} & ARC-Challenge & Acc. & 90.02 & \textbf{96.33} & 95.48 & 88.14 & 95.73 \\
 & GPQA-Diamond & Acc. & 76.77 & 75.76 & \textbf{79.29} & 54.04 & 67.68 \\
\midrule
Code & HumanEval+ & Pass@1 & \textbf{90.24} & 88.41 & 89.02 & 83.54 & 87.80 \\
\bottomrule
\end{tabular}
\end{table*}

\paragraph{Competitive Analysis.}
As illustrated in Table~\ref{tab:cross_model_new}, while Qwen3 variants exhibit robust performance on standard benchmarks (GSM8K, ARC-Challenge), Pangu-7B differentiates itself through superior proficiency in high-difficulty reasoning tasks:

\begin{itemize}
    \item \textbf{Math:} Pangu-7B attains an accuracy of 86.67\% on AIME-24, surpassing both Qwen3-8B (80.00\%) and the specialized DeepSeek-R1-Distill-Qwen3-8B (80.00\%) by a substantial margin. This suggests enhanced robustness in handling competition-level mathematical problems.

    \item \textbf{Code:} On the more stringent HumanEval+ benchmark, Pangu-7B reaches 90.24\%, outperforming Qwen3-14B (89.02\%) and leading the 8B-class models significantly.

    \item \textbf{General:} In expert-level GPQA-Diamond, Pangu-7B (76.77\%) exceeds its direct competitors Qwen3-8B (75.76\%) and DeepSeek-R1 variants, trailing only the larger Qwen3-14B model.
\end{itemize}

These findings imply that Pangu-7B's architecture trades marginal regressions in standard tasks for considerable gains in complex reasoning and synthesis capabilities, positioning it as a highly specialized model for demanding domains.


\subsubsection{Impact of Inference strategies}

We scrutinize the efficacy of two inference strategies defined in Sec.~\ref{sec:single_agent_reason}. Table~\ref{tab:pangu_foundational} tabulates the comparative results between the direct response (\texttt{no\_think}) and the adaptive reasoning (\texttt{auto\_think}) strategy.

\begin{table*}[t]
\centering
\normalsize
\setlength{\tabcolsep}{5pt}
\caption{Performance comparison of Pangu-7B under the two inference strategies.}
\label{tab:pangu_foundational}
\begin{tabular}{ll c ccc ccc}
\toprule
\multirow{2}{*}{\textbf{Domain}} & \multirow{2}{*}{\textbf{Task}} & \multirow{2}{*}{\textbf{Total}} & \multicolumn{2}{c}{\textbf{no\_think}} & & \multicolumn{3}{c}{\textbf{auto\_think}} \\
\cmidrule(lr){4-5} \cmidrule(lr){7-9}
 & & & Correct & Success Rate & & Correct & Success Rate & $\Delta$ \\
\midrule
\multirow{4}{*}{Math} 
 & AQUA & 254 & 223 & 87.80 & & 230 & \textbf{90.55} & \textcolor{blue}{+2.75} \\
 & GSM8K & 1319 & 1234 & 93.56 & & 1247 & \textbf{94.54} & \textcolor{blue}{+0.98} \\
 & GSM-Hard & 1319 & 814 & 61.71 & & 869 & \textbf{65.88} & \textcolor{blue}{+4.17} \\
 & AIME-2024 & 30 & 18 & 60.00 & & 26 & \textbf{86.67} & \textcolor{blue}{+26.67} \\
\midrule
\multirow{3}{*}{General} 
 & ARC-Easy & 2376 & 2233 & 93.98 & & 2281 & \textbf{96.00} & \textcolor{blue}{+2.02} \\
 & ARC-Challenge & 1172 & 1018 & 86.86 & & 1055 & \textbf{90.02} & \textcolor{blue}{+3.16} \\
 & GPQA-Diamond & 198 & 136 & 68.69 & & 152 & \textbf{76.77} & \textcolor{blue}{+8.08} \\
\midrule
\multirow{2}{*}{Code} 
 & HumanEval & 164 & 138 & 84.15 & & 157 & \textbf{95.73} & \textcolor{blue}{+11.58} \\
 & HumanEval+ & 164 & 130 & 79.27 & & 148 & \textbf{90.24} & \textcolor{blue}{+10.97} \\
\bottomrule
\end{tabular}
\end{table*}

Data presented in Table~\ref{tab:pangu_foundational} indicate that activating the \texttt{auto\_think} mechanism confers consistent performance uplifts. Specifically, these gains are most pronounced in frontier-level tasks necessitating complex logic synthesis, such as AIME-2024 (+26.67\%) and GPQA-Diamond (+8.08\%). This validates that the CoT procedure effectively bridges the gap between intuitive retrieval and rigorous problem-solving.

\subsection{Multi-Agent Inference Results}
\label{sec:mas_results}
We evaluate Pangu-7B under MAS workflows in Sec.~\ref{sec:mas_workflows} and compare them against Single-Model Inference, with results summarized in Table~\ref{tab:mas_no_think}.

Under the \texttt{no\_think} strategy, MAS workflows exhibit highly task-dependent effects. Reflection consistently improves performance across benchmarks, indicating strong self-correction capability, while Plan-and-Execute is particularly effective for structured tasks such as code generation. However, these gains come with clear trade-offs: strategies that benefit one task can impair others. For example, rigid Plan-and-Execution negatively impacts commonsense reasoning, and Adversarial Debate introduces substantial interference on tasks requiring precise, convergent logic. Overall, these results suggest that no single MAS design is universally optimal; effective collaboration patterns must be aligned with task characteristics.

We further examine MAS performance on top of (\texttt{auto\_think}) strategy, as shown in Table~\ref{tab:mas_auto_think}. While \texttt{auto\_think} substantially strengthens the baseline, additional multi-agent interactions provide limited and inconsistent benefits. In some cases, external debate complements internal reasoning, but in others, MAS integration leads to diminished performance. This pattern indicates diminishing returns during inference: once high-quality solutions are produced internally, additional agent interactions may introduce noise rather than useful evidence. This behavior is further illustrated through qualitative case studies in the Appendix~\ref{case study}.

\begin{table}[t]
\centering
\small
\caption{MAS results under the \texttt{no\_think} strategy. The delta ($\Delta$) compares accuracy to the single-model inference baseline shown in Table~\ref{tab:pangu_foundational}. The best-performing framework for each task is highlighted in bold.}
\label{tab:mas_no_think}
\setlength{\tabcolsep}{5pt}
\begin{tabular}{@{}llcc@{}}
\toprule
\textbf{Task} & \textbf{Selected MAS} & \textbf{Success Rate} & \textbf{$\Delta$} \\
\midrule
\multirow{4}{*}{GSM-Hard}      
& Plan-and-Execute        & 63.38 & \textcolor{blue}{+1.67} \\
& Interactive Debate      & 62.09 & \textcolor{blue}{+0.38} \\
& Reflection              & \textbf{67.78} & \textbf{\textcolor{blue}{+6.07}} \\
& Adversarial Debate      & 48.37 & \textcolor{blue}{-13.34} \\
\midrule
\multirow{4}{*}{ARC-Challenge} 
& Plan-and-Execute        & 79.35 & \textcolor{blue}{-7.51} \\
& Interactive Debate      & 90.27 & \textcolor{blue}{+3.41} \\
& Reflection              & \textbf{91.81} & \textbf{\textcolor{blue}{+4.95}} \\
& Adversarial Debate      & 82.51 & \textcolor{blue}{-4.35} \\
\midrule
\multirow{4}{*}{HumanEval} 
& Plan-and-Execute        & \textbf{92.68} & \textbf{\textcolor{blue}{+8.53}} \\
& Interactive Debate      & 85.37 & \textcolor{blue}{+1.22} \\
& Reflection              & 89.02 & \textcolor{blue}{+4.87} \\
& Adversarial Debate      & 78.66 & \textcolor{blue}{-5.49} \\
\bottomrule
\end{tabular}
\end{table}

\begin{table}[t]
\centering
\small
\caption{MAS results under the \texttt{auto\_think} strategy. The delta ($\Delta$) compares accuracy to the single-model inference baseline shown in Table~\ref{tab:pangu_foundational}.}
\label{tab:mas_auto_think}
\setlength{\tabcolsep}{6pt}
\begin{tabular}{@{}llcc@{}}
\toprule
\textbf{Task} & \textbf{Selected MAS} & \textbf{Success Rate} & \textbf{$\Delta$} \\
\midrule
GSM-Hard        & Plan-and-Execute    & 64.52 & \textcolor{blue}{-1.36} \\
ARC-Challenge  & Interactive Debate  & 91.55 & \textcolor{blue}{+1.53} \\
HumanEval      & Reflection          & 91.46 & \textcolor{blue}{-4.27} \\
\bottomrule
\end{tabular}
\end{table}

\subsection{Results on MIMeBench}
\label{sec:mimebench_results}

To assess the foundational skills of \textbf{semantic abstraction} and \textbf{contrastive discrimination}, we evaluated several 7B-scale models on MIMeBench, including general-purpose baselines (e.g., Qwen2.5-7B\footnote{\textbf{Qwen2.5-7B-Instruct}.~\url{https://www.modelscope.cn/models/Qwen/Qwen2.5-7B-Instruct}}, DeepSeek-7B\footnote{\textbf{DeepSeek-R1-Distill-Qwen-7B}.~\url{https://www.modelscope.cn/models/deepseek-ai/DeepSeek-R1-Distill-Qwen-7B}}) and a non-publicly available Specialized MCQ Generator. This analysis moves beyond final-answer correctness to assess the quality of the reasoning components themselves: identifying a main idea (abstraction) and constructing plausible yet incorrect alternatives (discrimination).

The results in Table~\ref{tab:qg_results_with_delta} reveal that proficiency in these foundational skills correlates with strong reasoning performance. While Pangu-7B underperforms the Specialized MCQ Generator, it demonstrates a clear advantage over other general-purpose baselines. This advantage is twofold:
\begin{itemize}
    \item First, Pangu-7B attains the highest correct-option score, a direct measure of its superior \textbf{semantic abstraction} capability in extracting a passage's central theme.
    \item Second, it generates the most effective distractors, evidenced by the highest mean distractor score. This indicates a stronger capacity for \textbf{contrastive discrimination}---the ability to create semantically challenging alternatives that test for true comprehension.
\end{itemize}

This direct evidence on foundational skills provides a compelling explanation for the robust performance Pangu-7B demonstrated on complex reasoning benchmarks(results in Sec.~\ref{sec:foundational_results}). A model that excels at identifying a problem's core semantics and distinguishing between nuanced, competing hypotheses is inherently better equipped to execute a reliable step-by-step reasoning process. The strength observed here is not merely about generating plausible text, but about the underlying \emph{semantic precision} that makes complex reasoning possible.

\begin{table}[t]
\centering
\caption{Evaluation results on MIMeBench. 
\textbf{Avg.} denotes the average dataset-level score $\overline{S}$ aggregated over all options; 
\textbf{Corr.} denotes the mean score $S^\star$ of the correct main-idea option; 
\textbf{Wrong.} denotes the mean score $S^-$ of three distractor options. 
\textbf{Specialized MCQ Generator} refers to a closed-source model adapted for MCQ generation.
All models are evaluated under the thinking strategy.}
\label{tab:qg_results_with_delta}
\begin{tabular}{@{}lccc@{}}
\toprule
\textbf{Model} & \textbf{Avg.} & \textbf{Corr.} & \textbf{Wrong.} \\
\midrule
Qwen2.5-7B     & 22.20 & 5.65 & 5.52 \\
Pangu-7B       & \underline{24.26} & \underline{6.38} & \underline{5.96} \\
DeepSeek-7B    & 21.53 & 5.25 & 5.43 \\
\textbf{Specialized MCQ Generator} & \textbf{28.08} & \textbf{8.97} & \textbf{6.37} \\
\bottomrule
\end{tabular}
\end{table}


%% file: sec/6_deepe.tex
\section{MAS Analysis: Roles, Cost, and Accuracy}
\label{sec:deeper-analy}

To complement the aggregate results reported in Section~\ref{sec:resul}, this section presents additional analyses that go beyond end-to-end accuracy. Specifically, we examine role-specific capability demands in MAS and analyze the trade-offs between inference cost and accuracy across different workflows.

\subsection{Role-Specific Capability Demand Analysis}
To better understand the capability demands imposed by different agent roles, we analyze model outputs under role-isolated MAS workflows.
Rather than focusing on the overall outcome of a MAS workflow, this analysis examines how individual roles---\emph{planner}, \emph{reviser}, and \emph{aggregator}---differ in the types of reasoning competence they require from a model.

For each MAS, the collaborative context is fixed and the evaluated model is assigned to a single role at a time (see Appendix~\ref{role-eval-append} for more details).
This allows different models to be compared under identical role-specific inputs, isolating how effectively they satisfy the capability requirements of each role, independent of interaction effects.

As shown in Table~\ref{tab:role_comparison}, the capability demands of different roles vary substantially.
Performance differences across models are relatively small for the \emph{Planner} and \emph{Aggregator} roles while the \emph{Reviser} role exhibits much larger variance.
Notably, Pangu-7B demonstrates a clear advantage in the Reviser role, achieving the highest revision accuracy among all compared models.
This suggests that its strength lies in post-hoc reasoning behaviors, including critiquing partially correct solutions and producing focused improvements, rather than in planning or aggregation alone.
Such results aligns with its strong performance under Reflection-based workflows observed in earlier experiments.

More broadly, this role-dependent pattern helps explain the heterogeneous effects observed in full multi-agent evaluations.
Workflows that hinge on revision or correction are more sensitive to reviser competence, whereas workflows centered on planning or aggregation are less discriminative with respect to model choice.
Overall, the analysis indicates that different agent roles place uneven demands on model capabilities, and that role-aware evaluation is necessary for interpreting multi-agent performance beyond aggregate accuracy.

\begin{table}[t]
  \centering
  \caption{Role-specific performance comparison under controlled role-isolation settings. Reviser is evaluated on HumanEval, Aggregator on ARC-Challenge, and Planner on GSM-Hard.}
  \label{tab:role_comparison}
  \normalsize
  \setlength{\tabcolsep}{6pt}
  \begin{tabular}{lccc}
    \toprule
    \textbf{Model} 
    & \textbf{Reviser}
    & \textbf{Aggregator} 
    & \textbf{Planner} \\
    \midrule
    Pangu-7B      & \textbf{92.07} & 91.38 & 68.69 \\
    Qwen2.5-7B    & 79.27          & 92.15 & \textbf{69.07} \\
    DeepSeek-7B   & 86.59          & \textbf{92.58} & 67.93 \\
    \bottomrule
  \end{tabular}
\end{table}

\subsection{Inference Cost and Accuracy Trade-offs}
We analyze the cost--accuracy trade-offs of different MAS workflows under the \texttt{no\_think} strategy, using ARC-Challenge as a representative benchmark, with total token consumption serving as a proxy for inference cost (Sec.~\ref{sec:multi_agent_reason}). While we focus on ARC-Challenge here, analogous analyses on additional benchmarks are reported in Appendix~\ref{cost-analysis} and exhibit consistent qualitative trends.

\begin{figure}[t]
  \centering
  \includegraphics[width=0.95\linewidth]{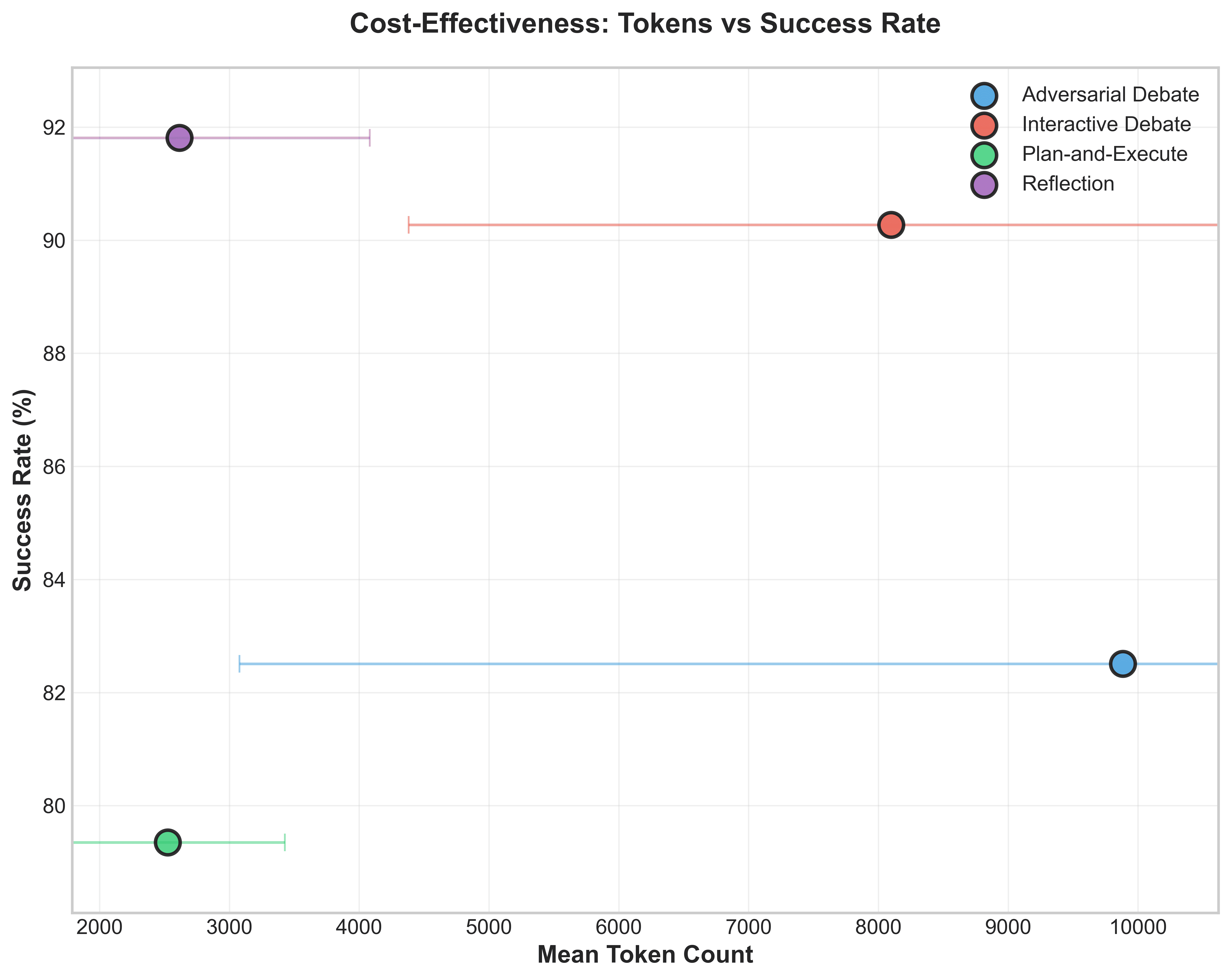}
  \caption{Mean token cost and success rate across MAS workflows on ARC-Challenge.}
  \label{fig:cost_arc_1}
\end{figure}

Fig.~\ref{fig:cost_arc_1} summarizes the overall cost-effectiveness frontier: \emph{Reflection} achieves the highest success rate while maintaining a low mean token cost, indicating that lightweight post-hoc correction can yield substantial quality gains without triggering large context growth. \emph{Interactive Debate} attains a comparable success rate but at a much higher average cost, suggesting diminishing returns when additional interaction rounds primarily add redundancy rather than decisive evidence. In contrast, \emph{Adversarial Debate} has the highest mean token cost while achieving only a mid-tier success rate, substantially trailing \emph{Reflection} and \emph{Interactive Debate}. Its extremely wide token range further suggests highly variable compute demand across instances, weakening its practical cost–reliability profile. \emph{Plan-and-Execute} operates at a similarly low token budget to \emph{Reflection}, but yields the lowest success rate among all methods on ARC-Challenge, indicating that the added structure does not translate into competitive accuracy in this setting.

\begin{figure}[t]
  \centering
  \includegraphics[width=0.95\linewidth]{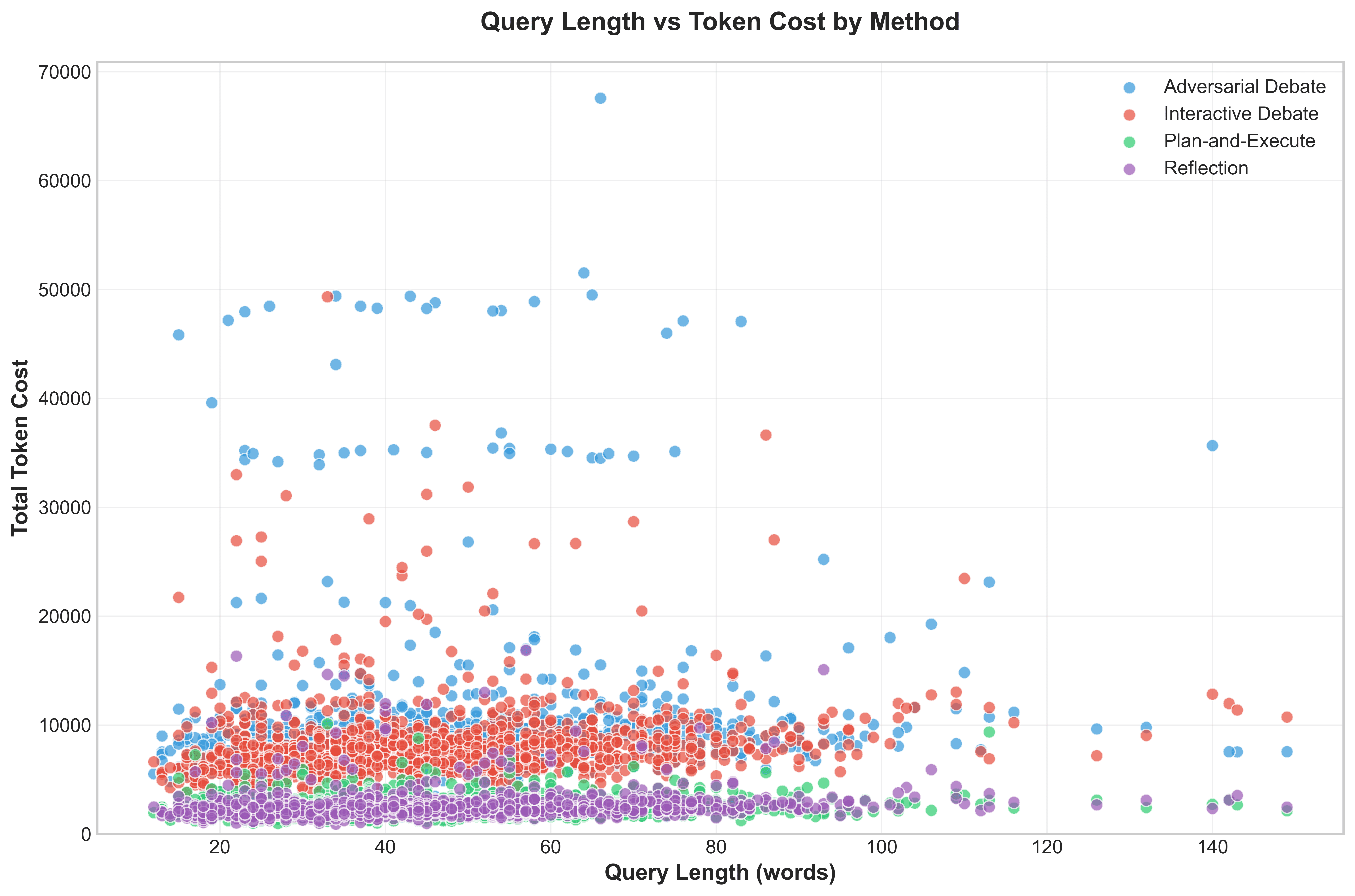}
  \caption{Query length versus total token cost for different MAS workflows on ARC-Challenge.}
  \label{fig:cost_arc_2}
\end{figure}

Fig.~\ref{fig:cost_arc_2} reveals that token cost is only weakly explained by input length. For all methods, token usage exhibits substantial dispersion at similar query lengths, implying that the dominant driver of cost is \emph{strategy-induced interaction dynamics} (e.g., number of turns, verbosity cascades, and transcript accumulation) rather than the query itself. Notably, debate-style methods exhibit a clear heavy-tail regime: a subset of instances triggers extremely long generations (up to $\sim 7\times 10^4$ tokens), reflecting a practical risk of cost blow-up under adversarial or multi-party exchanges. By comparison, \emph{Reflection} shows a much tighter band with limited outliers, indicating better cost controllability.

\begin{figure}[t]
  \centering
  \includegraphics[width=0.95\linewidth]{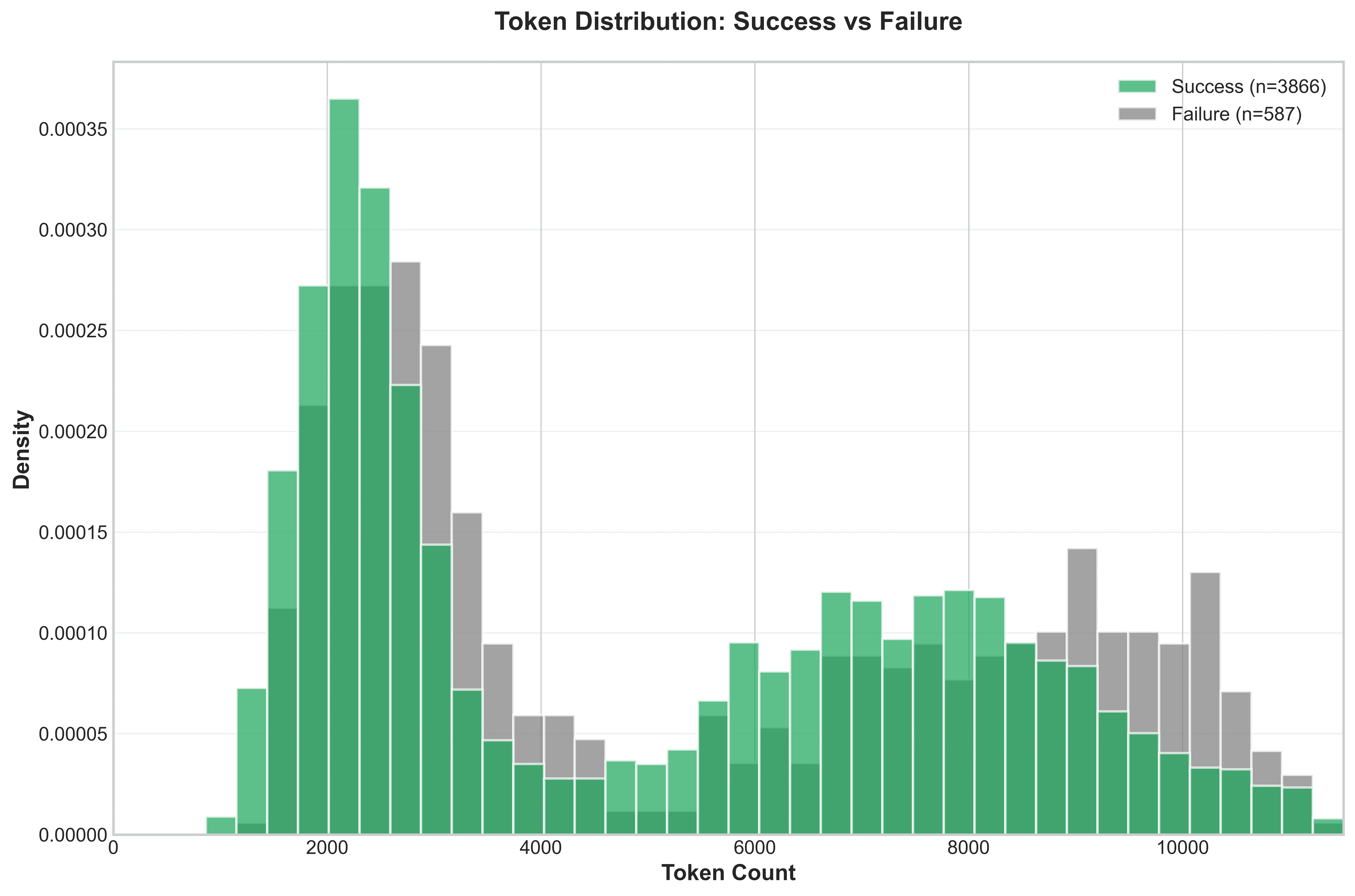}
  \caption{Token cost distributions for successful and failed instances on ARC-Challenge.}
  \label{fig:cost_arc_3}
\end{figure}

Finally, Fig.~\ref{fig:cost_arc_3} analyzes inference cost at the instance level. The token distribution is clearly bimodal, with a low-cost mode (roughly 2--4K tokens) and a high-cost mode (around 6--10K tokens), revealing substantial heterogeneity across problem instances. Crucially, failed cases are heavily concentrated in the high-cost regime. This indicates that elevated token consumption is not a signal of additional reasoning paying off, but rather a manifestation of the model struggling on inherently difficult instances--where more computation is expended without resolving the underlying uncertainty.

%% file: sec/7_concl.tex
\section{Conclusion}
\label{sec:conclusion}

This work presents a comprehensive investigation into the landscape of reasoning paradigms for LLMs, spanning from direct single-model generation and CoT augmentation to representative MAS. Our analysis reveals a critical trade-off: increased structural complexity does not guarantee improved reasoning. 
By evaluating these paradigms within a unified framework---integrating closed-form benchmarks with the novel evaluation axis introduced by our MIMeBench---we clarify the circumstances under which structural complexity provides meaningful improvements, as opposed to cases where it yields limited or unstable gains.
Ultimately, our findings provide a principled guide for the design and deployment of LLM-based reasoning systems, clarifying the intricate relationship between paradigm choice, performance reliability, and operational efficiency.

However, our study has several limitations. The analysis is mainly conducted on Pangu-7B model and a limited set of representative workflows, and the extent to which these findings generalize to other architectures or agent designs remains an open question. In addition, inference efficiency is primarily measured by token usage, which does not fully capture system-level latency or hardware constraints. Future work will extend this investigation to a broader range of models, and incorporate more comprehensive efficiency metrics to strengthen the empirical grounding of these findings.

%% file: sec/8_Appendix.tex
\clearpage
\setcounter{page}{1}

\appendix
\renewcommand{\thesection}{\Alph{section}} 
\setcounter{section}{0}

\section*{Appendix}
\addcontentsline{toc}{section}{Appendix} 

\section{Role-Isolation Evaluation Protocol}
\label{role-eval-append}

All experiments are conducted using open-weight models, with Pangu-7B, Qwen2.5-7B, and DeepSeek-7B evaluated under identical role-isolation settings.
For each target role, the evaluated model is substituted into that role while all other components of the multi-agent workflow are held fixed, enabling controlled comparison across models.

\begin{figure*}[t]
  \centering
  \includegraphics[width=0.95\textwidth]{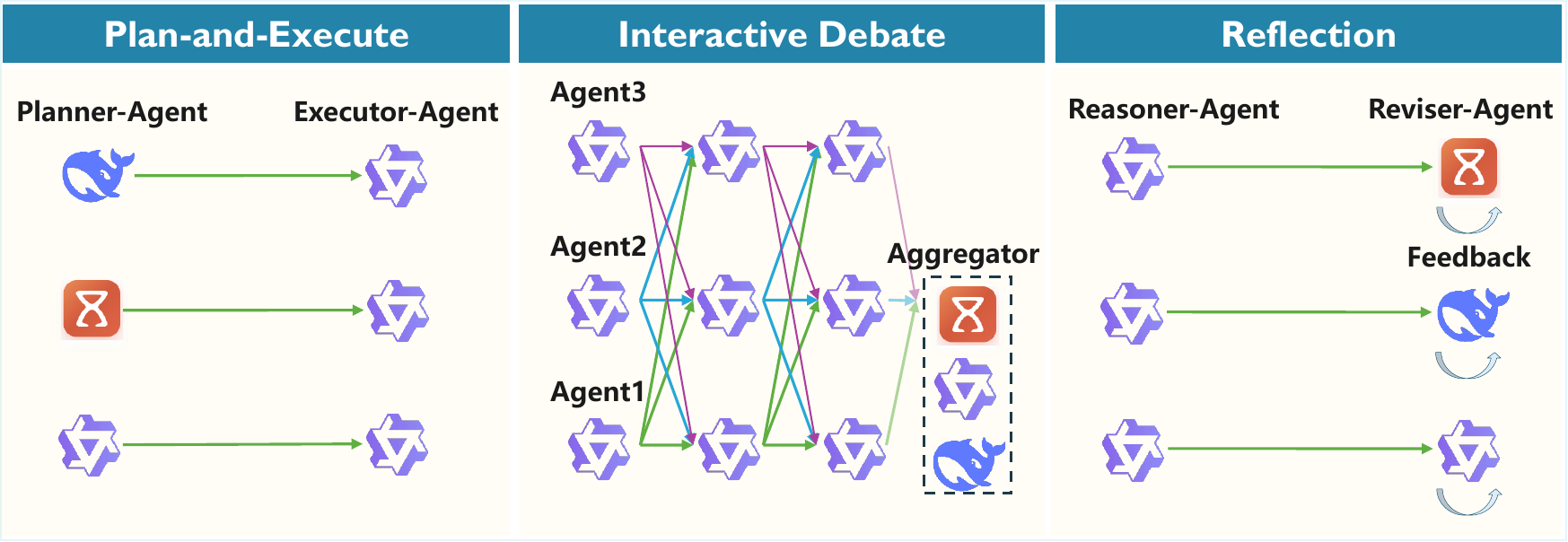}
  \caption{Role-Isolated Evaluation Workflow for Multi-Agent Systems.}
  \label{fig:mas-role}
\end{figure*}

For MAS that involve intermediate reasoning artifacts, including Reflection and Interactive Debate, we adopt a fixed-context evaluation protocol.
Specifically, all intermediate outputs(initial solution and debate messages) corresponding to non-target roles are generated once by a reference model (Qwen2.5-7B) and cached.
The evaluated model is then applied only to the target role and operates solely on these fixed artifacts.
This design ensures that different models receive identical role-specific inputs, isolating role competence from variability introduced by multi-agent interactions.
An illustration of this role-isolation setup is shown in Fig.~\ref{fig:mas-role}.

For the Plan-and-Execute workflow, fixed intermediate artifacts are not used, as the execution stage depends directly on the planner’s output.
Instead, to ensure fairness and reduce stochastic effects, the Executor is run with decoding temperature set to zero when evaluating planner-related behavior, so that execution differences are attributable solely to the planner’s output.

Across all role-isolation experiments, evaluation metrics and judging procedures are kept consistent with the main experiments.
For each role, performance is measured on a representative benchmark aligned with the role’s functional responsibility (HumanEval for Reviser, ARC-Challenge for Aggregator, and GSM-Hard for Planner), allowing focused and interpretable role-level comparison.

\section{Prompt Template}
\label{prompt}
\begin{figure}[t]
  \centering
  \includegraphics[width=0.95\linewidth]{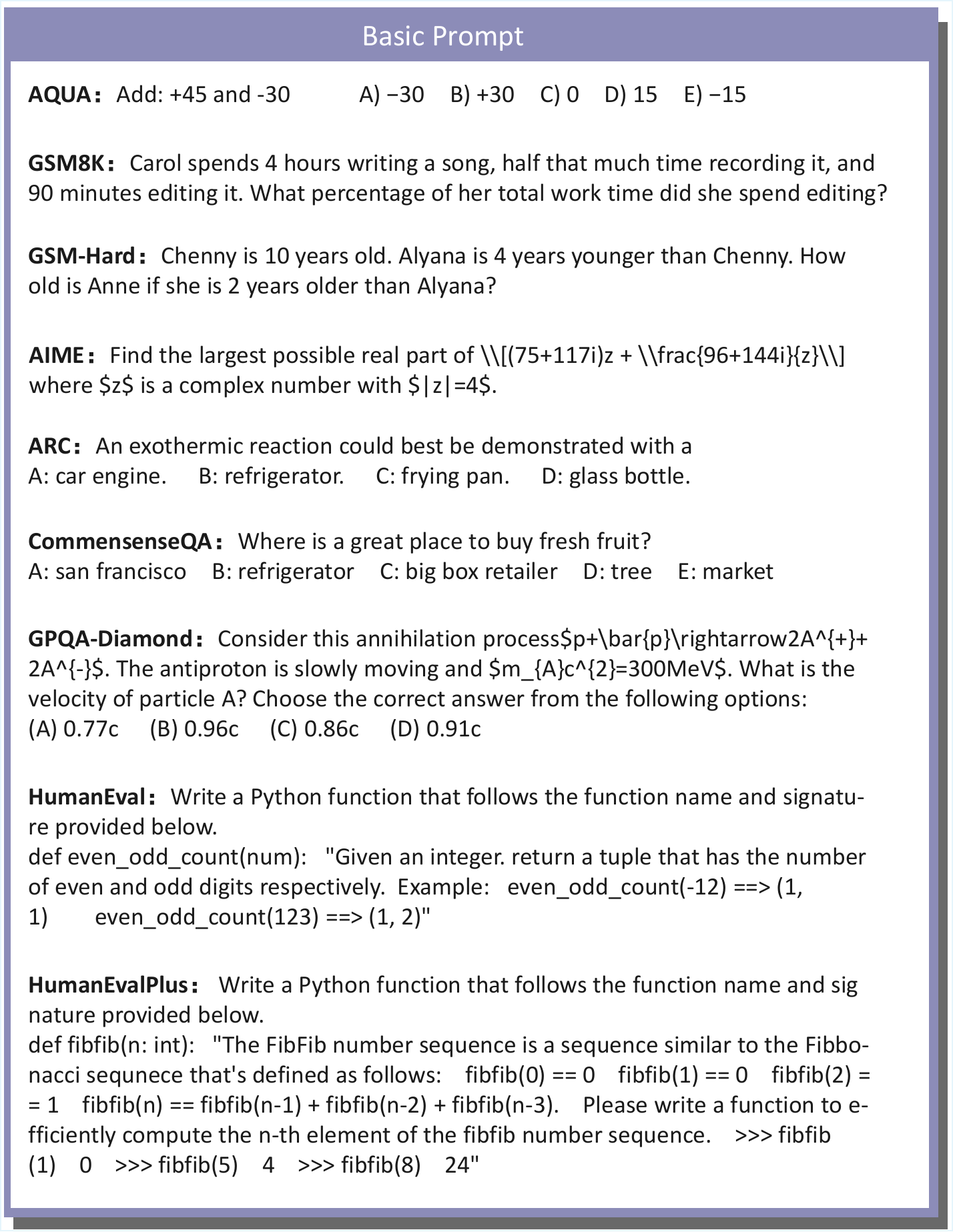}
  \caption{Basic prompt template (single-model inference).}
  \label{fig:prompt-1}
\end{figure}

\begin{figure}[t]
  \centering
  \includegraphics[width=0.95\linewidth]{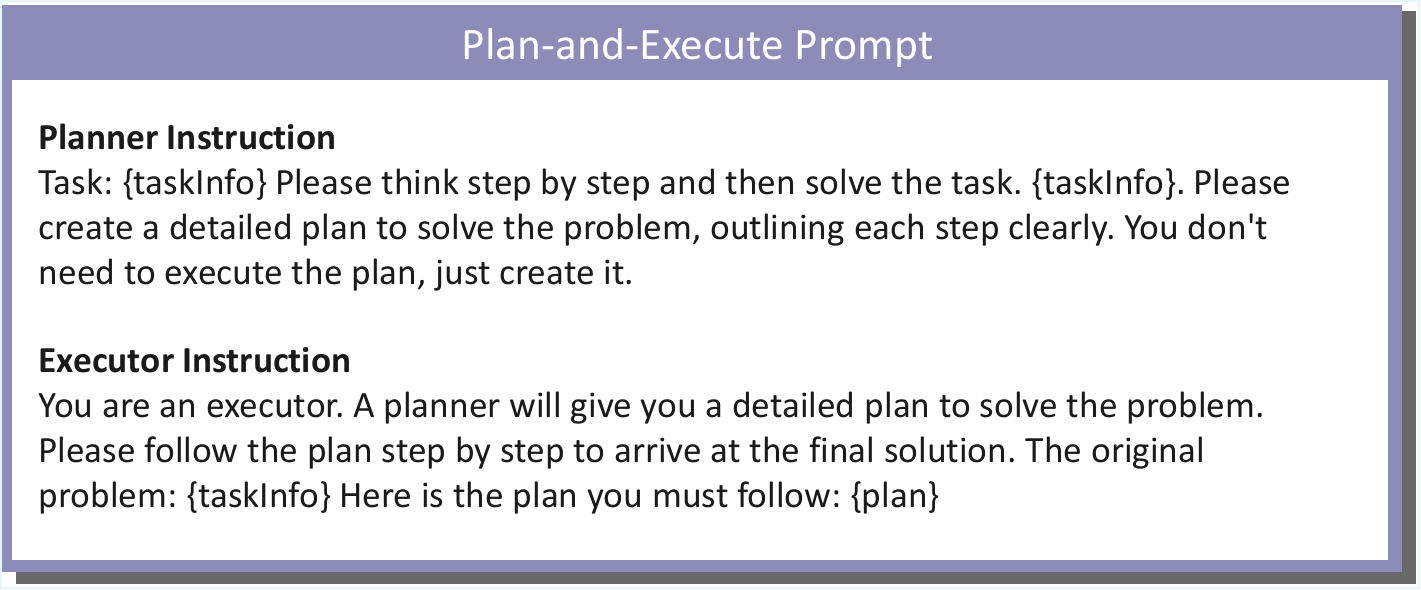}
  \caption{Plan-and-Execute prompt templates (Planner \& Executor).}
  \label{fig:prompt-2}
\end{figure}

\begin{figure}[t]
  \centering
  \includegraphics[width=0.95\linewidth]{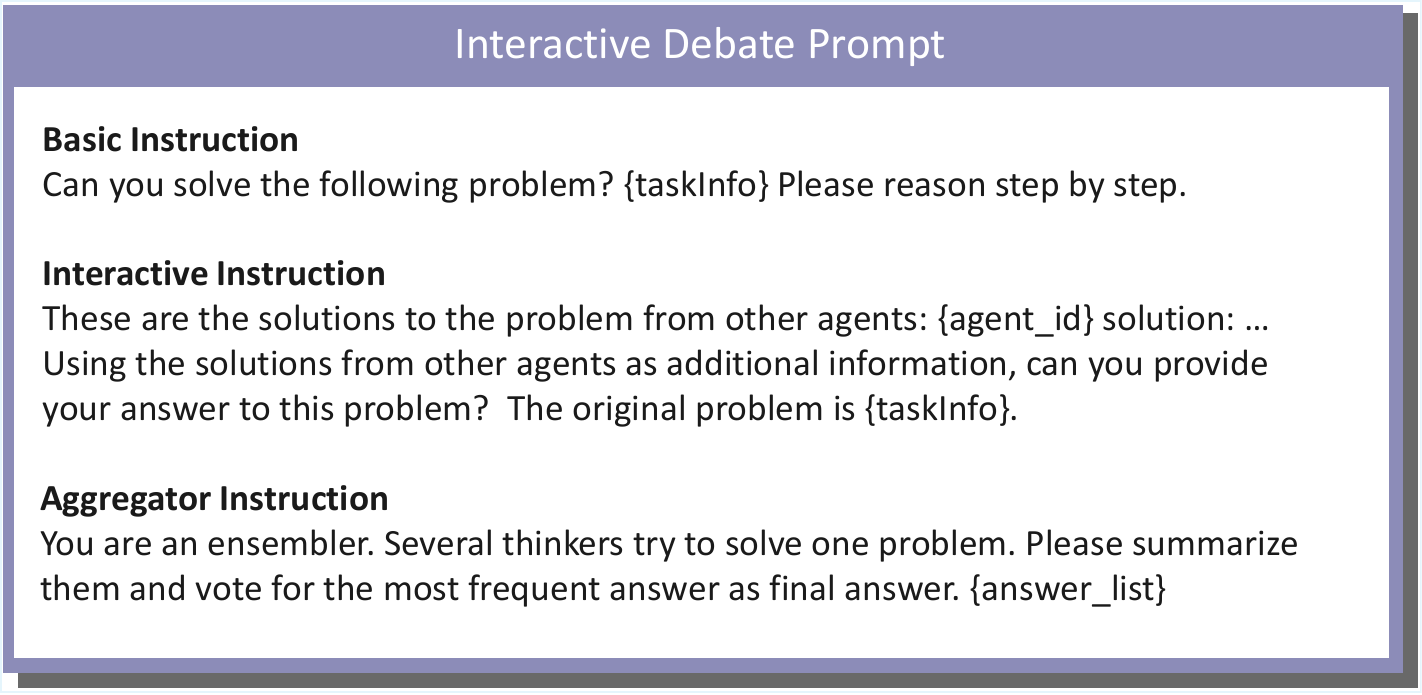}
  \caption{Interactive Debate prompt templates (Debaters \& Aggregator).}
  \label{fig:prompt-3}
\end{figure}

\begin{figure}[t]
  \centering
  \includegraphics[width=0.95\linewidth]{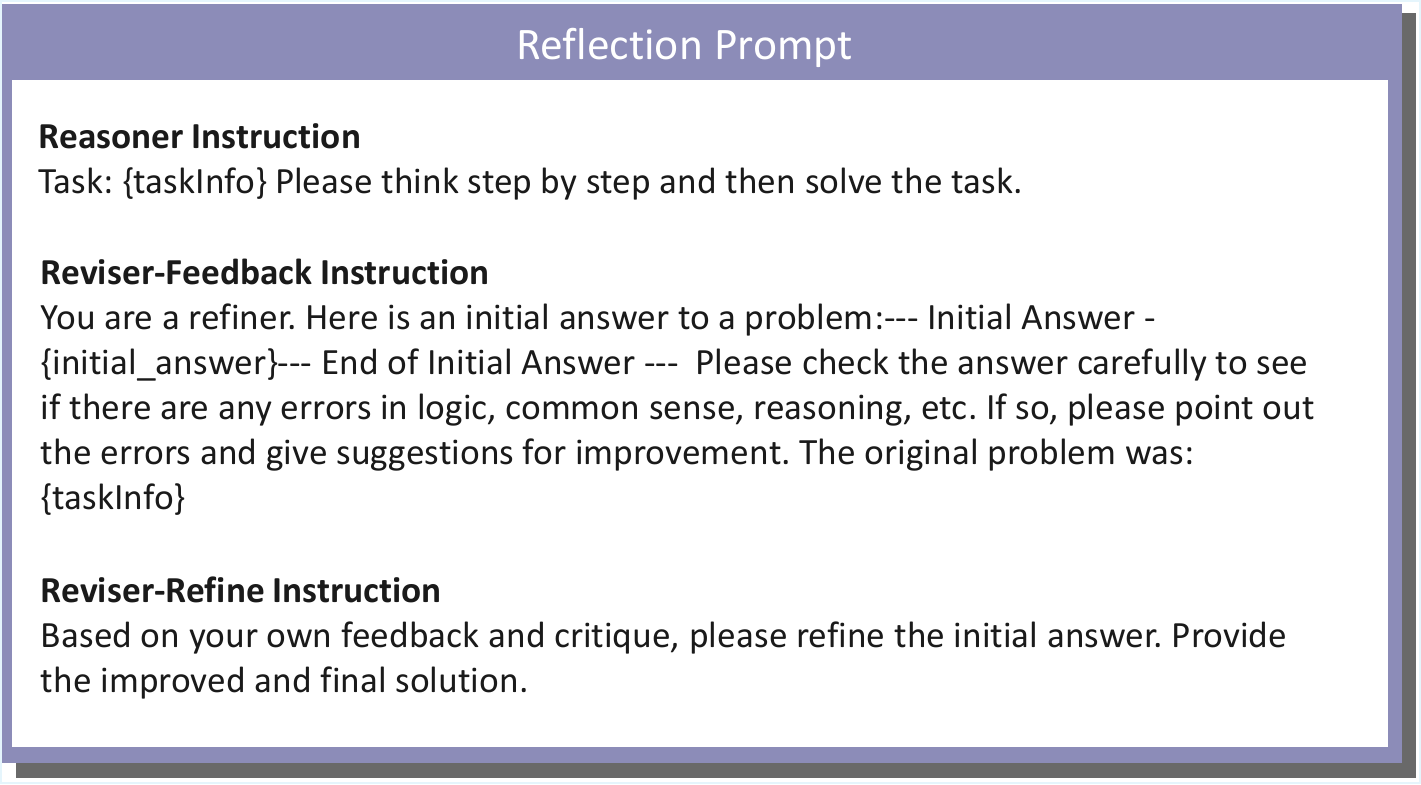}
  \caption{Reflection prompt templates (Reasoner \& Reviser).}
  \label{fig:prompt-4}
\end{figure}

\begin{figure}[t]
  \centering
  \includegraphics[width=0.95\linewidth]{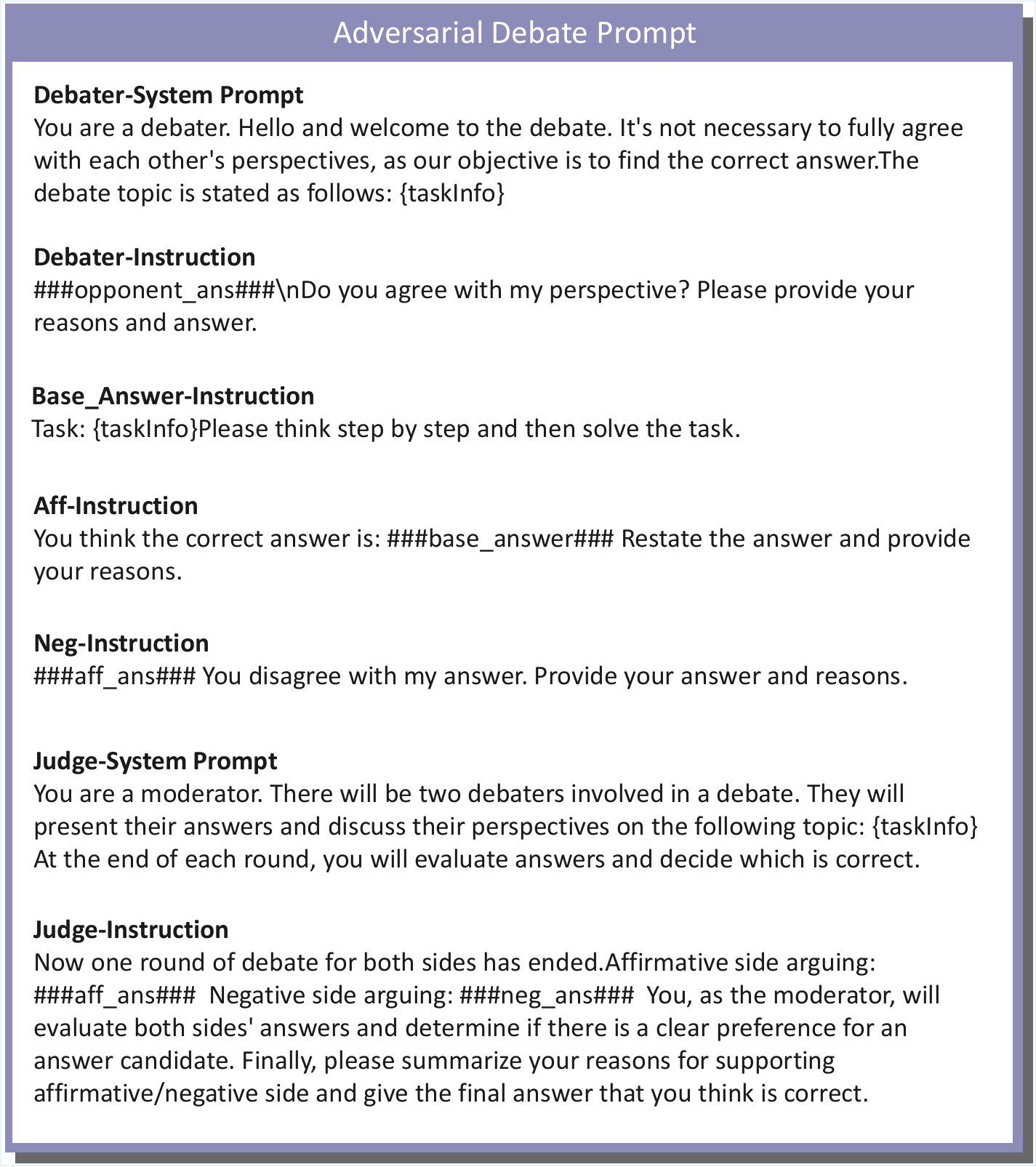}
  \caption{Adversarial Debate prompt templates (Affirmative, Negative, Judge).}
  \label{fig:prompt-5}
\end{figure}

\begin{figure}[t]
  \centering
  \includegraphics[width=0.95\linewidth]{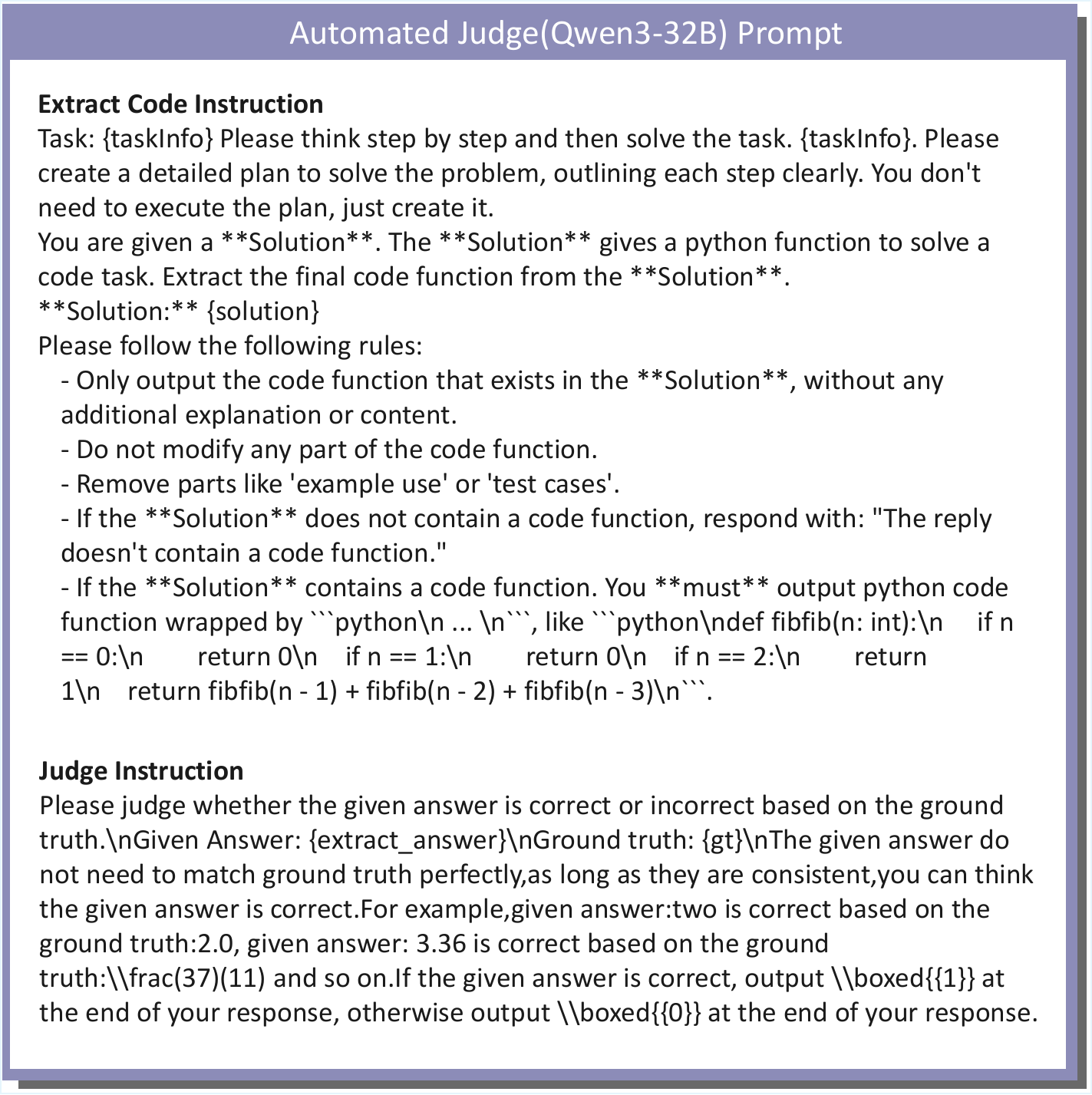}
  \caption{Automated judge prompt templates (Qwen3-32B) for evaluation.}
  \label{fig:prompt-6}
\end{figure}

\begin{figure}[t]
  \centering
  \includegraphics[width=\linewidth]{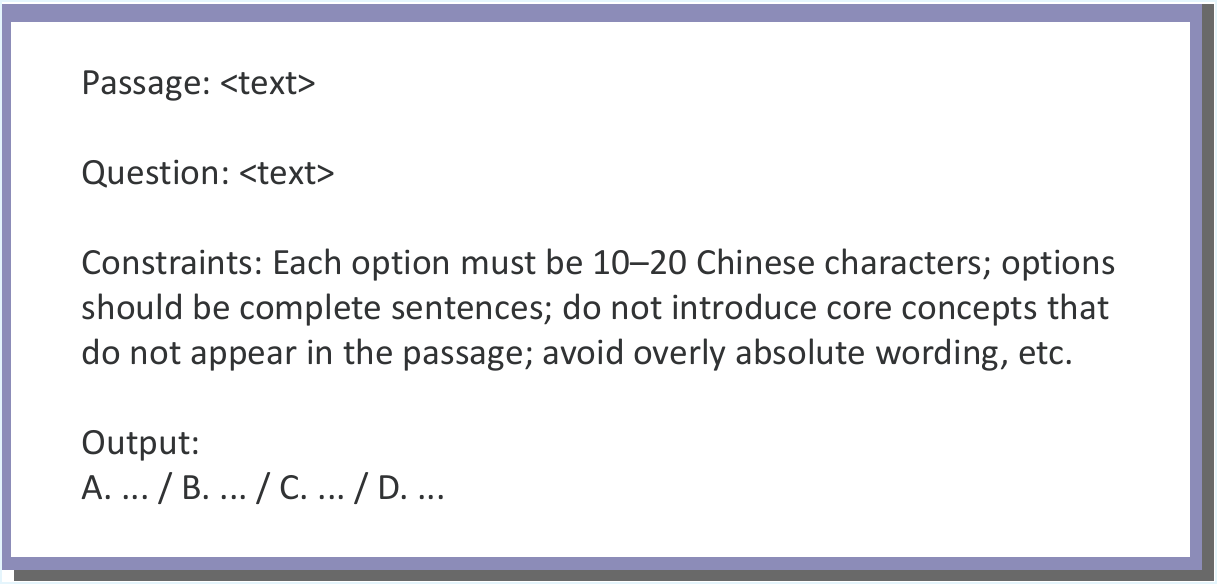}
  \caption{Illustrative prompt template for models evaluated on MIMeBench (fields only; no real content). Input includes the passage, question, optional constraints (e.g., length or format), and the required output format; the model should output four structured options (A--D) while satisfying length and style constraints.}
  \label{fig:prompt-7}
\end{figure}

Fig.~\ref{fig:prompt-1}--\ref{fig:prompt-7} show prompt templates of our study.

\section{Cost and Accuracy Analysis} 
\label{cost-analysis} 

\begin{figure}[t]
  \centering
  \includegraphics[width=0.95\linewidth]{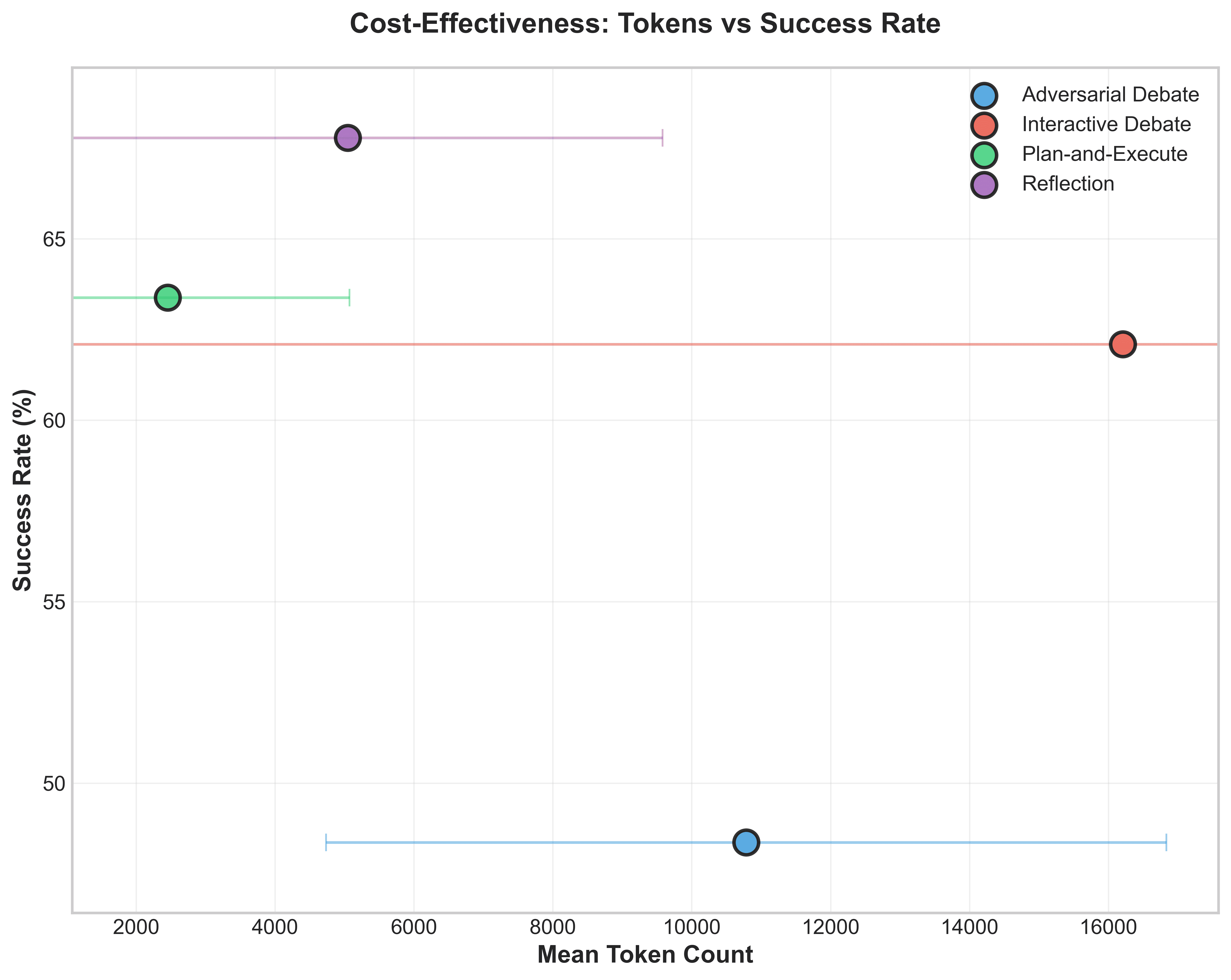}
  \caption{Mean token cost and success rate across MAS workflows on GSM-Hard.}
  \label{fig:cost_gsm_1}
\end{figure}

\begin{figure}[t]
  \centering
  \includegraphics[width=0.95\linewidth]{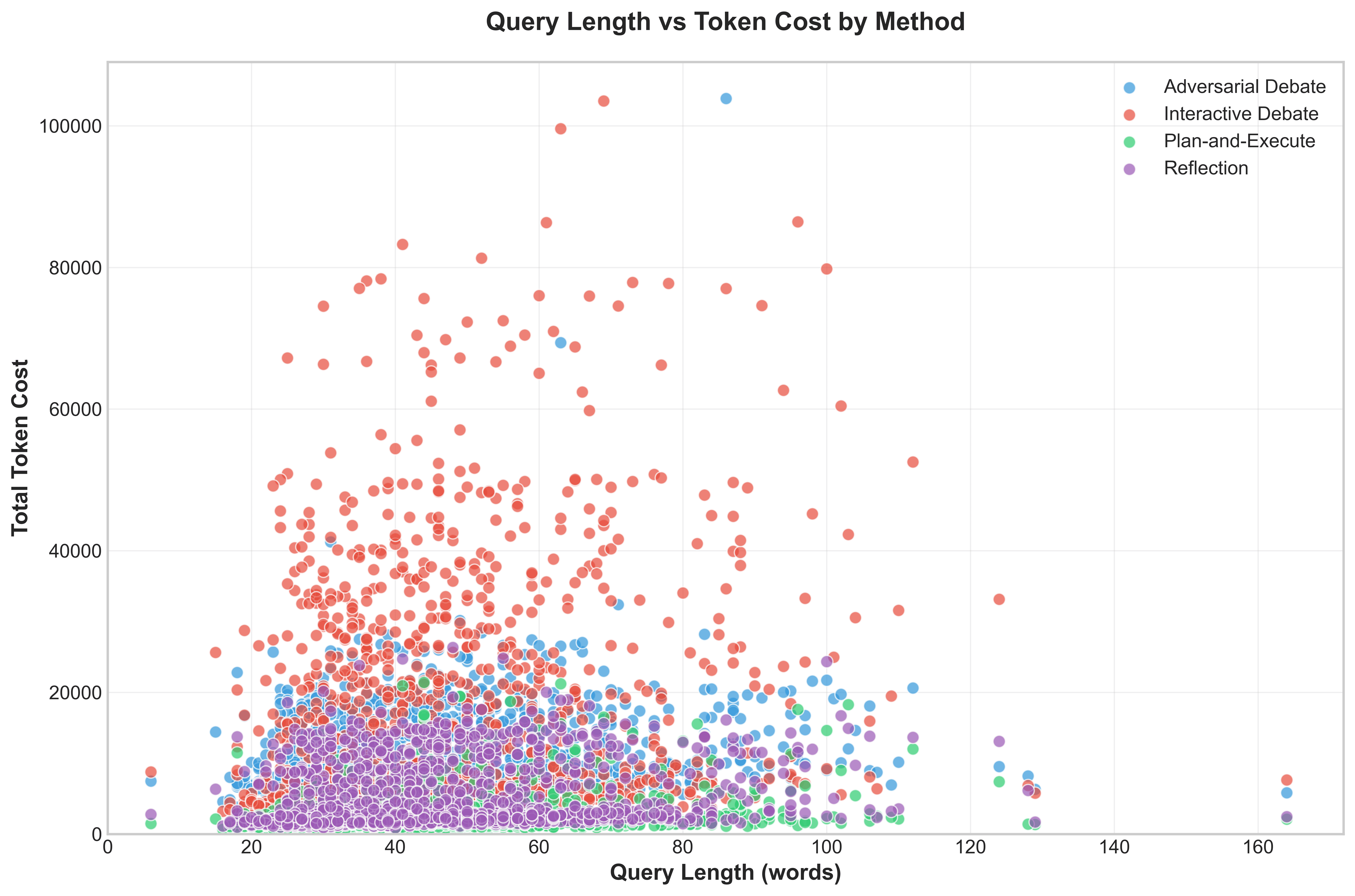}
  \caption{Query length versus total token cost for different MAS workflows on GSM-Hard.}
  \label{fig:cost_gsm_2}
\end{figure}

\begin{figure}[t]
  \centering
  \includegraphics[width=0.95\linewidth]{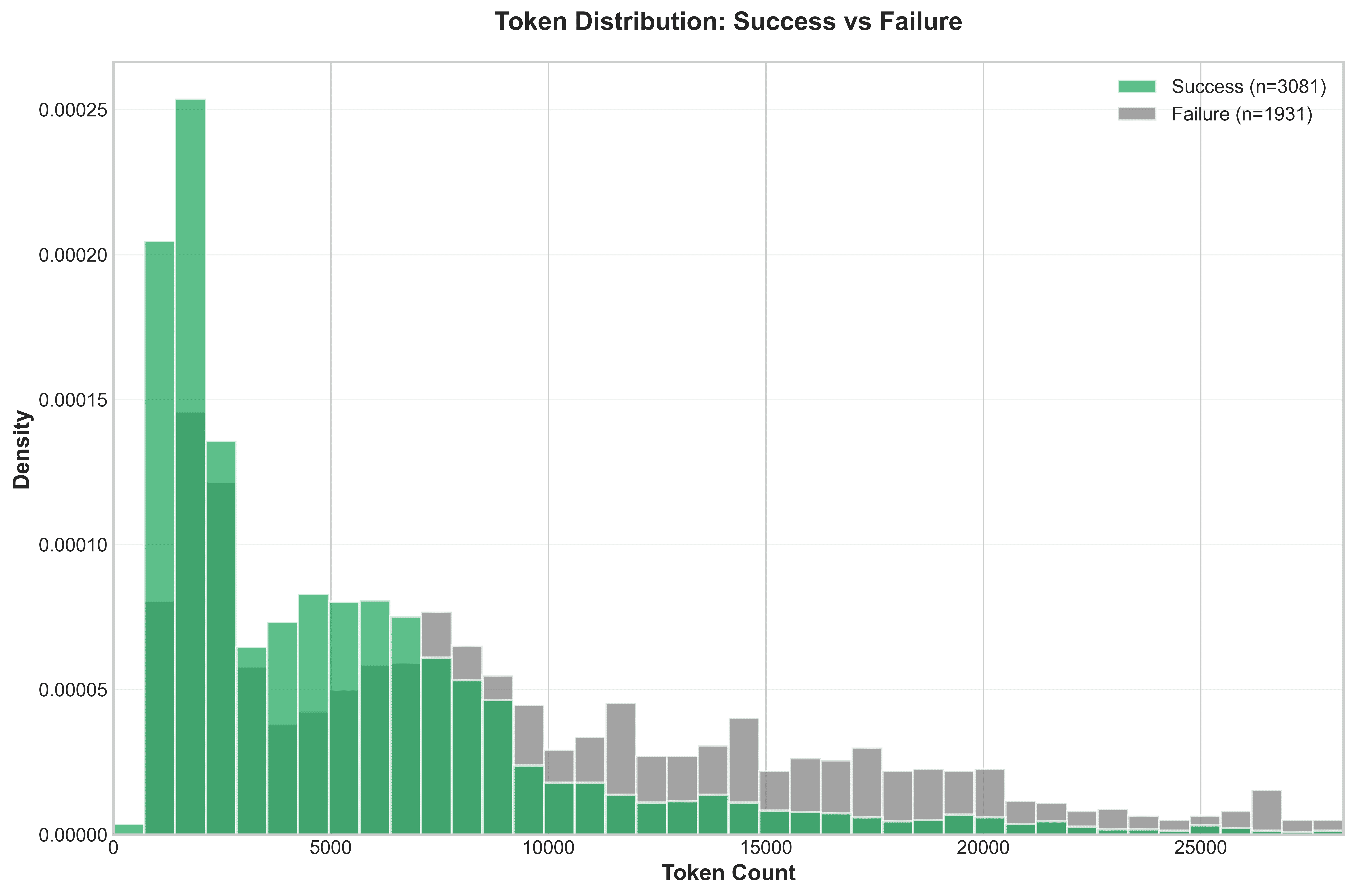}
  \caption{Token cost distributions for successful and failed instances on GSM-Hard.}
  \label{fig:cost_gsm_3}
\end{figure}

\begin{figure}[t]
  \centering
  \includegraphics[width=0.95\linewidth]{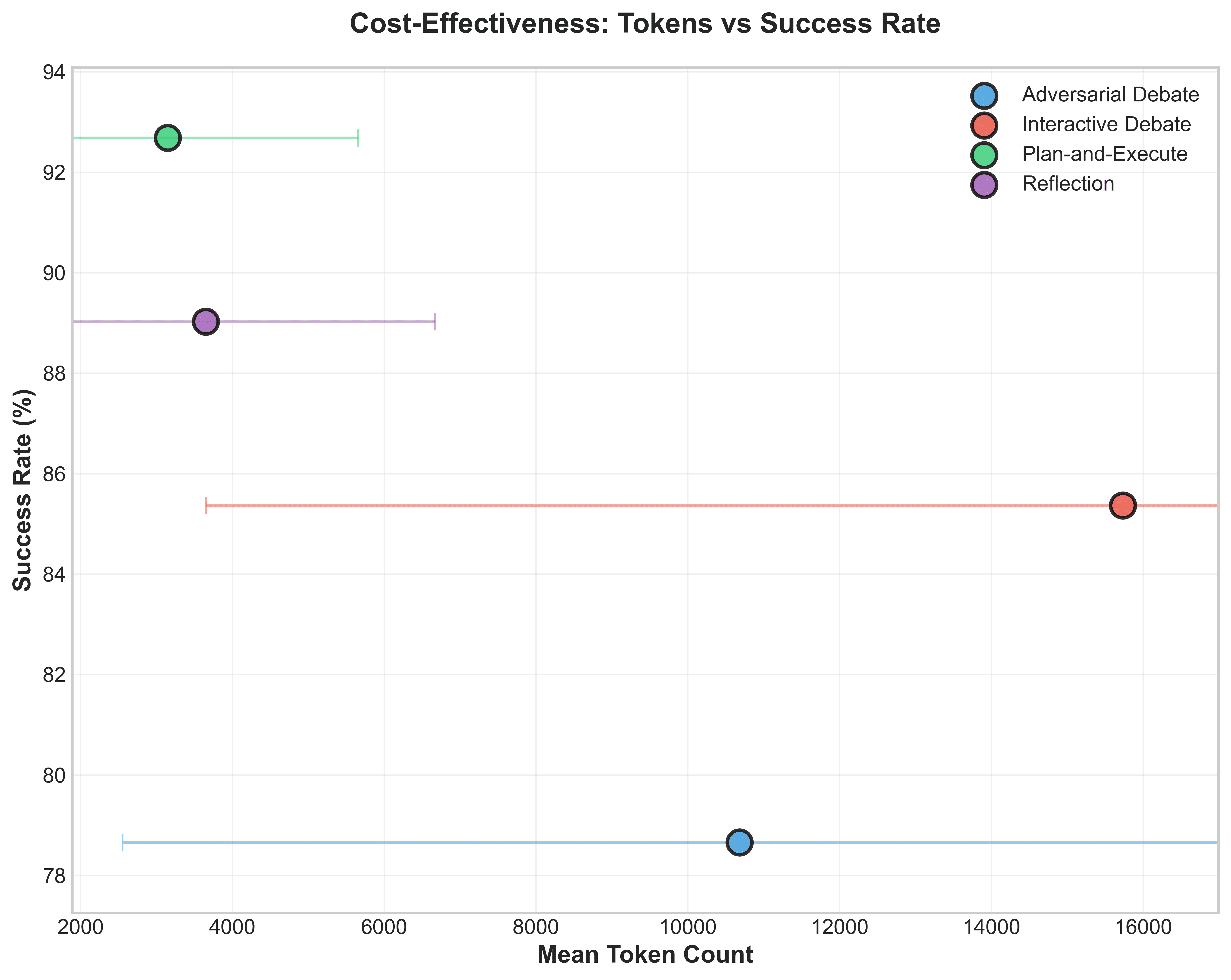}
  \caption{Mean token cost and success rate across MAS workflows on Humaneval.}
  \label{fig:cost_humaneval_1}
\end{figure}

\begin{figure}[t]
  \centering
  \includegraphics[width=0.95\linewidth]{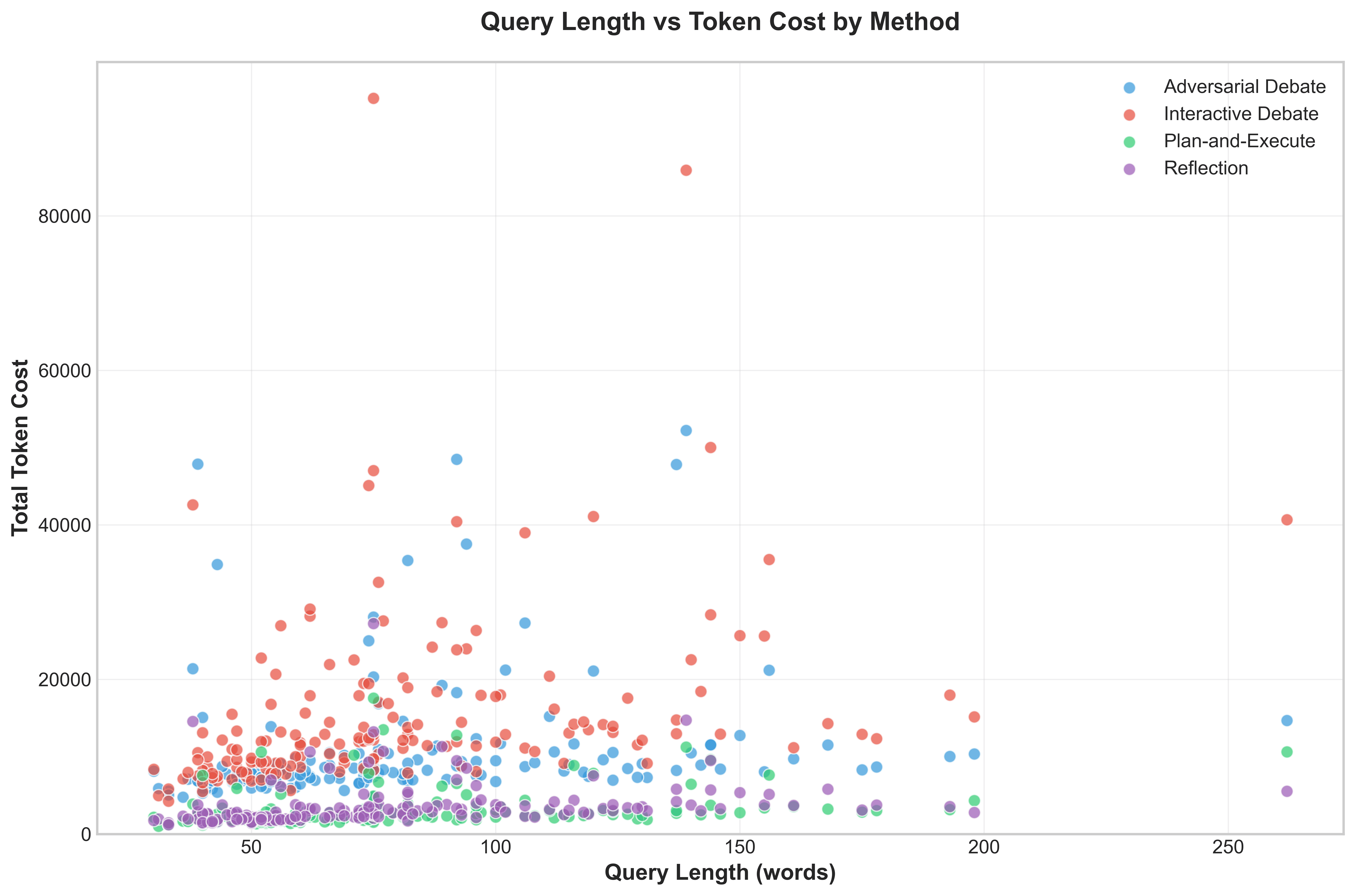}
  \caption{Query length versus total token cost for different MAS workflows on HumanEval.}
  \label{fig:cost_humaneval_2}
\end{figure}

\begin{figure}[t]
  \centering
  \includegraphics[width=0.95\linewidth]{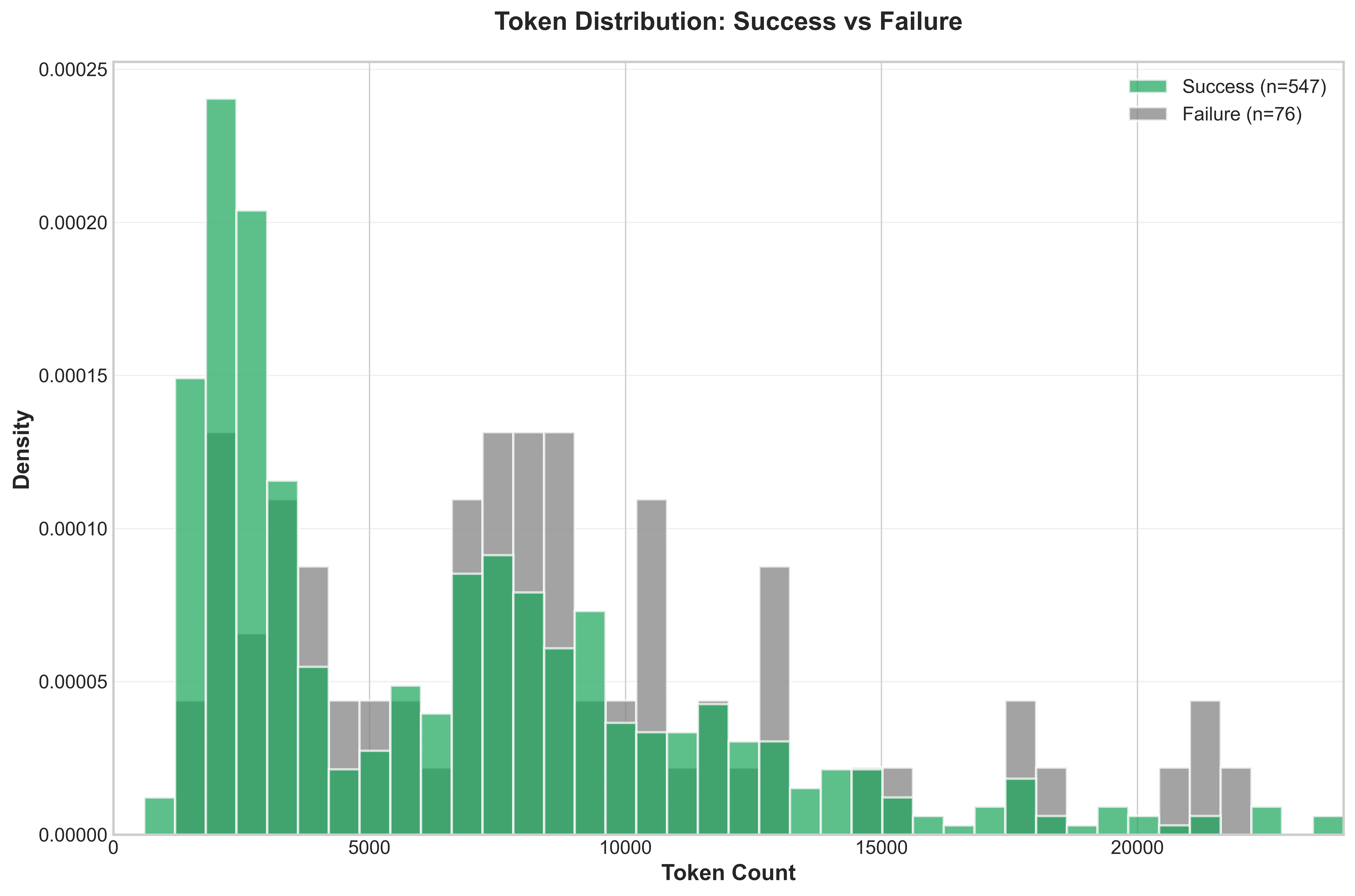}
  \caption{Token cost distributions for successful and failed instances on HumanEval.}
  \label{fig:cost_humaneval_3}
\end{figure}

As shown in Fig.~\ref{fig:cost_gsm_1}--\ref{fig:cost_gsm_3} for GSM-Hard and
Fig.~\ref{fig:cost_humaneval_1}--\ref{fig:cost_humaneval_3} for HumanEval,
both benchmarks exhibit trends that are qualitatively consistent with
those observed on ARC-Challenge.

\begin{figure}[t]
  \centering
  \includegraphics[width=0.95\linewidth]{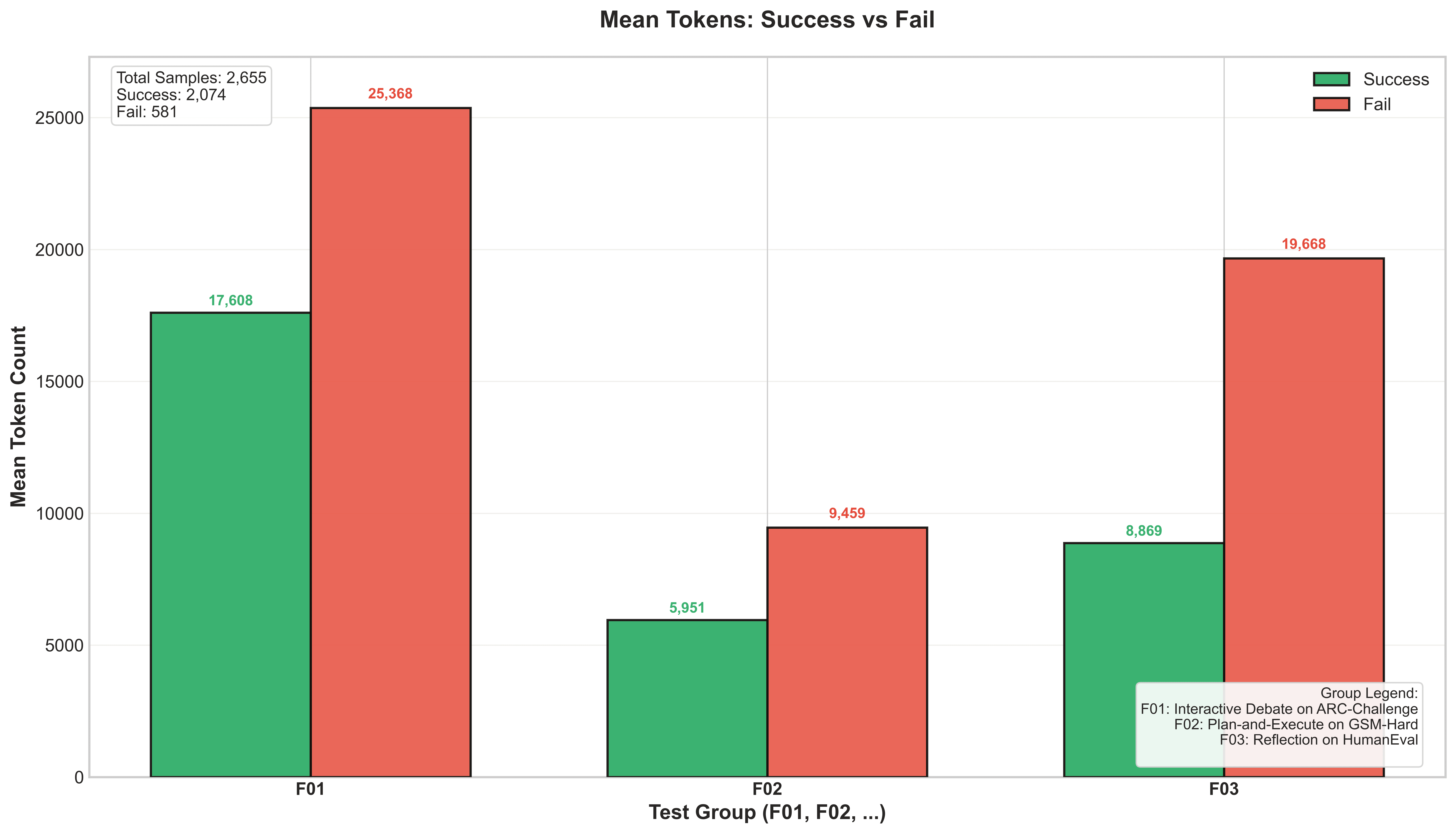}
  \caption{Mean token cost for successful and failed instances under \texttt{auto\_think} strategy.}
  \label{fig:cost_auto_think}
\end{figure}

In addition, Fig.~\ref{fig:cost_auto_think} serves as a supplementary cost analysis under the \texttt{auto-think} strategy, extending the no-think results presented earlier. Consistent with previous observations, failed instances consume substantially more tokens than successful ones across all workflows, indicating that higher inference cost remains associated with reasoning instability rather than improved outcomes, even when internal deliberation is enabled. This confirms that the cost–accuracy patterns identified under no-think persist under auto-think, reinforcing the robustness of our conclusions.

\section{Case Study} 
\label{case study}

\begin{figure}[t]
  \centering
  \includegraphics[width=0.95\linewidth]{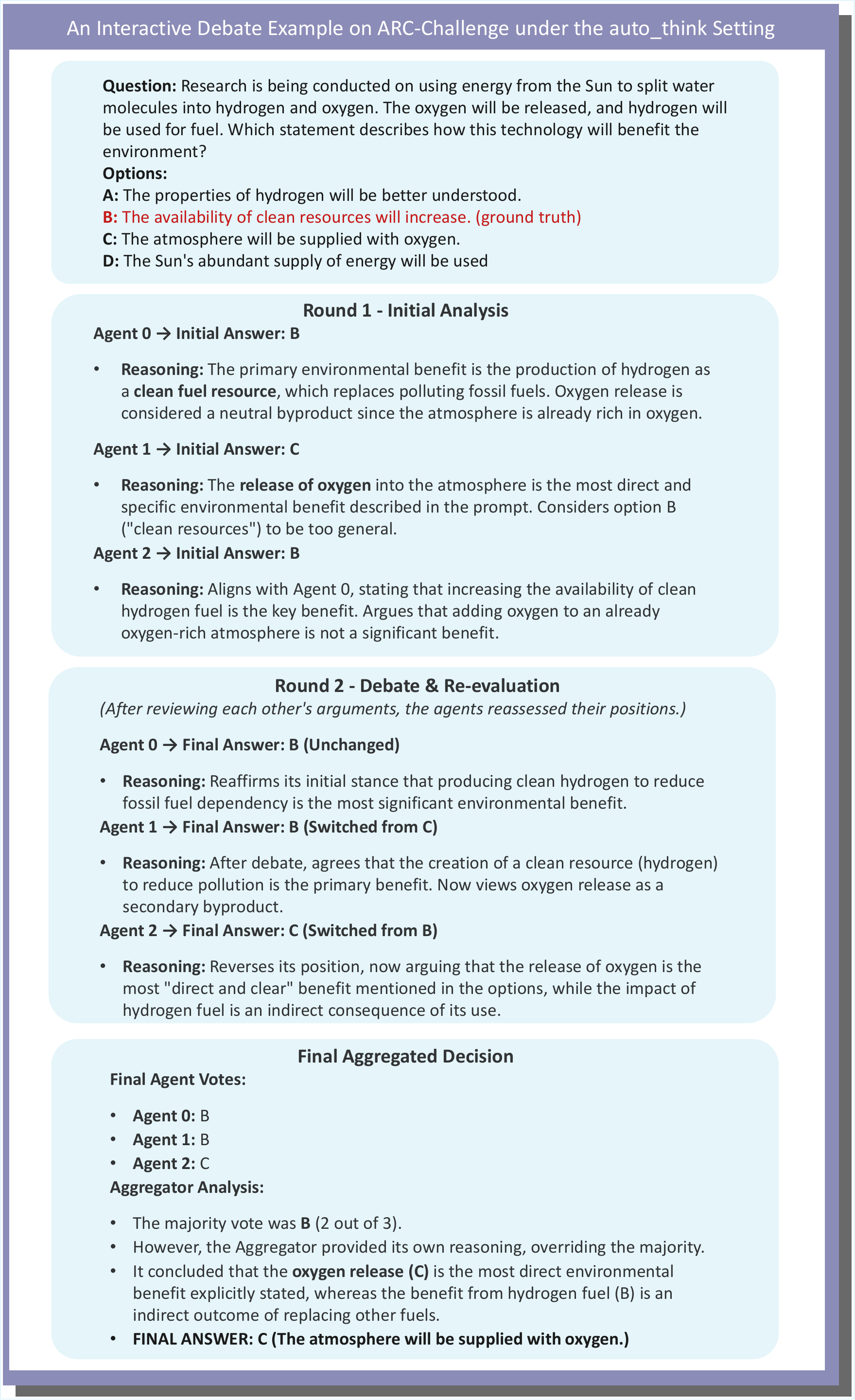}
  \caption{
An example of interactive debate on the ARC-Challenge task under the \texttt{auto\_think} strategy.
Three agents first generate independent answers by a CoT procedure, followed by a debate and re-evaluation phase.
While the majority of agents favor the clean-resource interpretation, the ensemble ultimately selects an alternative option based on explicit semantic alignment.
This case illustrates that, when a strong CoT procedure is already present, additional multi-agent interactions may lead to inconsistent outcomes and diminished returns.
}
  \label{fig:case-study}
\end{figure}

Figure~\ref{fig:case-study} provides a concrete example of the Interactive Debate process on ARC-Challenge under the \texttt{auto\_think} strategy, illustrating the results discussed in Table~\ref{tab:mas_auto_think}.

%% file: main.bbl
\begin{thebibliography}{32}
\providecommand{\natexlab}[1]{#1}
\providecommand{\url}[1]{\texttt{#1}}
\expandafter\ifx\csname urlstyle\endcsname\relax
  \providecommand{\doi}[1]{doi: #1}\else
  \providecommand{\doi}{doi: \begingroup \urlstyle{rm}\Url}\fi

\bibitem[Chen et~al.(2025)Chen, Wang, Han, Li, Li, Bi, Li, Wang, Mi, Zhu, et~al.]{chen2025pangu}
Hanting Chen, Yasheng Wang, Kai Han, Dong Li, Lin Li, Zhenni Bi, Jinpeng Li, Haoyu Wang, Fei Mi, Mingjian Zhu, et~al.
\newblock Pangu embedded: An efficient dual-system llm reasoner with metacognition.
\newblock \emph{arXiv preprint arXiv:2505.22375}, 2025.

\bibitem[Chen et~al.(2024{\natexlab{a}})Chen, Saha, and Bansal]{chen2024reconcile}
Justin Chen, Swarnadeep Saha, and Mohit Bansal.
\newblock Reconcile: Round-table conference improves reasoning via consensus among diverse llms.
\newblock In \emph{Proceedings of the 62nd Annual Meeting of the Association for Computational Linguistics (Volume 1: Long Papers)}, pages 7066--7085, 2024{\natexlab{a}}.

\bibitem[Chen et~al.(2021)Chen, Tworek, Jun, Yuan, de~Oliveira~Pinto, et~al.]{chen2021evaluating}
Mark Chen, Jerry Tworek, Heewoo Jun, Qiming Yuan, Henrique~Ponde de Oliveira~Pinto, et~al.
\newblock Evaluating large language models trained on code, 2021.

\bibitem[Chen et~al.(2024{\natexlab{b}})Chen, Su, Zuo, Yang, Yuan, Chan, Yu, Lu, Hung, Qian, et~al.]{chen2024agentverse}
Weize Chen, Yusheng Su, Jingwei Zuo, Cheng Yang, Chenfei Yuan, Chi-Min Chan, Heyang Yu, Yaxi Lu, Yi-Hsin Hung, Chen Qian, et~al.
\newblock Agentverse: Facilitating multi-agent collaboration and exploring emergent behaviors.
\newblock In \emph{ICLR}, 2024{\natexlab{b}}.

\bibitem[Clark et~al.(2018)Clark, Cowhey, Etzioni, Khot, Sabharwal, et~al.]{allenai:arc}
Peter Clark, Isaac Cowhey, Oren Etzioni, Tushar Khot, Ashish Sabharwal, et~al.
\newblock Think you have solved question answering? try arc, the ai2 reasoning challenge.
\newblock \emph{arXiv:1803.05457v1}, 2018.

\bibitem[Cobbe et~al.(2021)Cobbe, Kosaraju, Bavarian, Chen, Jun, et~al.]{cobbe2021gsm8k}
Karl Cobbe, Vineet Kosaraju, Mohammad Bavarian, Mark Chen, Heewoo Jun, et~al.
\newblock Training verifiers to solve math word problems.
\newblock \emph{arXiv preprint arXiv:2110.14168}, 2021.

\bibitem[DeepSeek-AI(2025)]{deepseekai2025deepseekr1incentivizingreasoningcapability}
DeepSeek-AI.
\newblock Deepseek-r1: Incentivizing reasoning capability in llms via reinforcement learning, 2025.

\bibitem[Du et~al.(2023)Du, Li, Torralba, Tenenbaum, and Mordatch]{du2023improving}
Yilun Du, Shuang Li, Antonio Torralba, Joshua~B Tenenbaum, and Igor Mordatch.
\newblock Improving factuality and reasoning in language models through multiagent debate.
\newblock In \emph{Forty-first International Conference on Machine Learning}, 2023.

\bibitem[Fu et~al.(2024)Fu, Ng, Jiang, and Liu]{fu2024gptscore}
Jinlan Fu, See~Kiong Ng, Zhengbao Jiang, and Pengfei Liu.
\newblock Gptscore: Evaluate as you desire.
\newblock In \emph{Proceedings of the 2024 Conference of the North American Chapter of the Association for Computational Linguistics: Human Language Technologies (Volume 1: Long Papers)}, pages 6556--6576, 2024.

\bibitem[Gao et~al.(2022)Gao, Madaan, Zhou, Alon, Liu, et~al.]{gao2022pal}
Luyu Gao, Aman Madaan, Shuyan Zhou, Uri Alon, Pengfei Liu, et~al.
\newblock Pal: Program-aided language models.
\newblock \emph{arXiv preprint arXiv:2211.10435}, 2022.

\bibitem[Hong et~al.(2023)Hong, Zhuge, Chen, Zheng, Cheng, Wang, Zhang, Wang, Yau, Lin, et~al.]{hong2023metagpt}
Sirui Hong, Mingchen Zhuge, Jonathan Chen, Xiawu Zheng, Yuheng Cheng, Jinlin Wang, Ceyao Zhang, Zili Wang, Steven Ka~Shing Yau, Zijuan Lin, et~al.
\newblock Metagpt: Meta programming for a multi-agent collaborative framework.
\newblock In \emph{The Twelfth International Conference on Learning Representations}, 2023.

\bibitem[Kumar(2024)]{kumar2024large}
Pranjal Kumar.
\newblock Large language models (llms): survey, technical frameworks, and future challenges.
\newblock \emph{Artificial Intelligence Review}, 57\penalty0 (10):\penalty0 260, 2024.

\bibitem[Li et~al.(2024)Li, Wang, Zeng, Wu, and Yang]{li2024survey}
Xinyi Li, Sai Wang, Siqi Zeng, Yu Wu, and Yi Yang.
\newblock A survey on llm-based multi-agent systems: workflow, infrastructure, and challenges.
\newblock \emph{Vicinagearth}, 1\penalty0 (1):\penalty0 9, 2024.

\bibitem[Liang et~al.(2024)Liang, He, Jiao, Wang, Wang, Wang, Yang, Shi, and Tu]{liang2024encouraging}
Tian Liang, Zhiwei He, Wenxiang Jiao, Xing Wang, Yan Wang, Rui Wang, Yujiu Yang, Shuming Shi, and Zhaopeng Tu.
\newblock Encouraging divergent thinking in large language models through multi-agent debate.
\newblock In \emph{Proceedings of the 2024 conference on empirical methods in natural language processing}, pages 17889--17904, 2024.

\bibitem[Lifshitz et~al.(2025)Lifshitz, McIlraith, and Du]{lifshitz2025multi}
Shalev Lifshitz, Sheila~A McIlraith, and Yilun Du.
\newblock Multi-agent verification: Scaling test-time compute with multiple verifiers.
\newblock \emph{arXiv preprint arXiv:2502.20379}, 2025.

\bibitem[Lin(2004)]{lin2004rouge}
Chin-Yew Lin.
\newblock Rouge: A package for automatic evaluation of summaries.
\newblock In \emph{Text summarization branches out}, pages 74--81, 2004.

\bibitem[Ling et~al.(2017)Ling, Yogatama, Dyer, and Blunsom]{ling2017program}
Wang Ling, Dani Yogatama, Chris Dyer, and Phil Blunsom.
\newblock Program induction by rationale generation: Learning to solve and explain algebraic word problems.
\newblock In \emph{Proceedings of the 55th Annual Meeting of the Association for Computational Linguistics (Volume 1: Long Papers)}, pages 158--167, 2017.

\bibitem[Liu et~al.(2023{\natexlab{a}})Liu, Xia, Wang, and Zhang]{evalplus}
Jiawei Liu, Chunqiu~Steven Xia, Yuyao Wang, and Lingming Zhang.
\newblock Is your code generated by chat{GPT} really correct? rigorous evaluation of large language models for code generation.
\newblock In \emph{Thirty-seventh Conference on Neural Information Processing Systems}, 2023{\natexlab{a}}.

\bibitem[Liu et~al.(2023{\natexlab{b}})Liu, Iter, Xu, Wang, Xu, and Zhu]{liu2023g}
Yang Liu, Dan Iter, Yichong Xu, Shuohang Wang, Ruochen Xu, and Chenguang Zhu.
\newblock G-eval: Nlg evaluation using gpt-4 with better human alignment.
\newblock \emph{arXiv preprint arXiv:2303.16634}, 2023{\natexlab{b}}.

\bibitem[Madaan et~al.(2023)Madaan, Tandon, Gupta, Hallinan, Gao, Wiegreffe, Alon, Dziri, Prabhumoye, Yang, et~al.]{madaan2023self}
Aman Madaan, Niket Tandon, Prakhar Gupta, Skyler Hallinan, Luyu Gao, Sarah Wiegreffe, Uri Alon, Nouha Dziri, Shrimai Prabhumoye, Yiming Yang, et~al.
\newblock Self-refine: Iterative refinement with self-feedback.
\newblock \emph{Advances in Neural Information Processing Systems}, 36:\penalty0 46534--46594, 2023.

\bibitem[Minaee et~al.(2024)Minaee, Mikolov, Nikzad, Chenaghlu, Socher, Amatriain, and Gao]{minaee2024large}
Shervin Minaee, Tomas Mikolov, Narjes Nikzad, Meysam Chenaghlu, Richard Socher, Xavier Amatriain, and Jianfeng Gao.
\newblock Large language models: A survey.
\newblock \emph{arXiv preprint arXiv:2402.06196}, 2024.

\bibitem[Rein et~al.(2024)Rein, Hou, Stickland, Petty, Pang, et~al.]{rein2024gpqa}
David Rein, Betty~Li Hou, Asa~Cooper Stickland, Jackson Petty, Richard~Yuanzhe Pang, et~al.
\newblock {GPQA}: A graduate-level google-proof q\&a benchmark.
\newblock In \emph{First Conference on Language Modeling}, 2024.

\bibitem[Renze and Guven(2024)]{renze2024self}
Matthew Renze and Erhan Guven.
\newblock Self-reflection in llm agents: Effects on problem-solving performance.
\newblock \emph{arXiv preprint arXiv:2405.06682}, 2024.

\bibitem[Shinn et~al.(2023)Shinn, Cassano, Gopinath, Narasimhan, and Yao]{shinn2023reflexion}
Noah Shinn, Federico Cassano, Ashwin Gopinath, Karthik Narasimhan, and Shunyu Yao.
\newblock Reflexion: Language agents with verbal reinforcement learning.
\newblock \emph{Advances in Neural Information Processing Systems}, 36:\penalty0 8634--8652, 2023.

\bibitem[Talmor et~al.(2019)Talmor, Herzig, Lourie, and Berant]{talmor-etal-2019-commonsenseqa}
Alon Talmor, Jonathan Herzig, Nicholas Lourie, and Jonathan Berant.
\newblock {C}ommonsense{QA}: A question answering challenge targeting commonsense knowledge.
\newblock In \emph{Proceedings of the 2019 Conference of the North {A}merican Chapter of the Association for Computational Linguistics: Human Language Technologies, Volume 1 (Long and Short Papers)}, pages 4149--4158, Minneapolis, Minnesota, 2019. Association for Computational Linguistics.

\bibitem[Team(2025)]{qwen3technicalreport}
Qwen Team.
\newblock Qwen3 technical report, 2025.

\bibitem[Tran et~al.(2025)Tran, Dao, Nguyen, Pham, O'Sullivan, and Nguyen]{tran2025multi}
Khanh-Tung Tran, Dung Dao, Minh-Duong Nguyen, Quoc-Viet Pham, Barry O'Sullivan, and Hoang~D Nguyen.
\newblock Multi-agent collaboration mechanisms: A survey of llms.
\newblock \emph{arXiv preprint arXiv:2501.06322}, 2025.

\bibitem[Wang et~al.()Wang, Wei, Schuurmans, Le, Chi, Narang, Chowdhery, and Zhou]{wangself}
Xuezhi Wang, Jason Wei, Dale Schuurmans, Quoc~V Le, Ed~H Chi, Sharan Narang, Aakanksha Chowdhery, and Denny Zhou.
\newblock Self-consistency improves chain of thought reasoning in language models.
\newblock In \emph{The Eleventh International Conference on Learning Representations}.

\bibitem[Wei et~al.(2022)Wei, Wang, Schuurmans, Bosma, Xia, et~al.]{wei2022chain}
Jason Wei, Xuezhi Wang, Dale Schuurmans, Maarten Bosma, Fei Xia, et~al.
\newblock Chain-of-thought prompting elicits reasoning in large language models.
\newblock \emph{Advances in neural information processing systems}, 35:\penalty0 24824--24837, 2022.

\bibitem[Yoran et~al.()Yoran, Wolfson, Bogin, Katz, Deutch, and Berant]{yoran2023answering}
Ori Yoran, Tomer Wolfson, Ben Bogin, Uri Katz, Daniel Deutch, and Jonathan Berant.
\newblock Answering questions by meta-reasoning over multiple chains of thought.
\newblock In \emph{The 2023 Conference on Empirical Methods in Natural Language Processing}.

\bibitem[Zhao et~al.(2023)Zhao, Zhou, Li, Tang, Wang, Hou, Min, Zhang, Zhang, Dong, et~al.]{zhao2023survey}
Wayne~Xin Zhao, Kun Zhou, Junyi Li, Tianyi Tang, Xiaolei Wang, Yupeng Hou, Yingqian Min, Beichen Zhang, Junjie Zhang, Zican Dong, et~al.
\newblock A survey of large language models.
\newblock \emph{arXiv preprint arXiv:2303.18223}, 1\penalty0 (2), 2023.

\bibitem[Zheng et~al.(2023)Zheng, Chiang, Sheng, Zhuang, Wu, Zhuang, Lin, Li, Li, Xing, et~al.]{zheng2023judging}
Lianmin Zheng, Wei-Lin Chiang, Ying Sheng, Siyuan Zhuang, Zhanghao Wu, Yonghao Zhuang, Zi Lin, Zhuohan Li, Dacheng Li, Eric Xing, et~al.
\newblock Judging llm-as-a-judge with mt-bench and chatbot arena.
\newblock \emph{Advances in neural information processing systems}, 36:\penalty0 46595--46623, 2023.

\end{thebibliography}
